# Distributed Intrusion Detection for the Security of Societies of Robots

Adriano Fagiolini, *Member, IEEE,* Gianluca Dini, and Antonio Bicchi, *Fellow, IEEE*

*Abstract*—This paper addresses the problem of detecting possible intruders in a group of autonomous robots, which coexist in a shared environment and interact with each other according to a set of "social behaviors", or common rules. Such rules specify what actions each robot is allowed to perform in the pursuit of its individual goals: rules are distributed, i.e. they can evaluated based only on the state of the individual robot, and on information that can be sensed directly or through communication with immediate neighbors.

We consider intruders as robots which misbehave, i.e. do not follow the rules, because of either spontaneous failures or malicious reprogramming. Our goal is to detect intruders by observing the congruence of their behavior with the social rules as applied to the current state of the overall system. Moreover, in accordance with the fully distributed nature of the problem, the detection itself must be peformed by individual robots, based only on local information.

The paper introduces a formalism that allows to model uniformly a large variety of possible robot societies. The main contribution consists in the proposal of an Intrusion Detection System, i.e. a protocol that, under suitabkle conditions, allows individual robots to detect possible misbehaving robots in their vicinity, and trigger possible further actions to secure the society.

The protocol is based on two main components. The first is a *monitor* that an individual robot runs, using only locally available information, and applies to each of its neighbors, which constructs the set of all possible system states which might explain the observed behavior and are consistent with its own, limited, direct knowlegde. The second component is a set–valued consensus algorithm allowing different "views" estimated by different local monitors to be combined through communication.

Sufficient conditions are given, which ensure finite–time convergence towards a consensus state, and hence towards the decision that would be made by a hypothetical centralized IDS with complete information. It is worth noting that the generality of the protocol formalism makes so that local monitors can be automatically generated once the cooperation rules and the robot dynamics are specified. The effectiveness of the proposed technique is shown through application to examples of automated robotic systems.

## I. Introduction

THAT the availability of distributed systems gave rise in the late 80s to a profound rethinking of many decision making problems [1]–[3], and even enabled solutions to them that could not be possible before [4], [5], is on everyone's records. That this is happening also in Control and will soon involve many applications in Robotics can easily be foreseen. Various distributed control policies have been proposed for e.g. formation control, flocking, sensor coverage, and intelligent transportation (see e.g. [6]–[9]). An intrinsic paradigm shift is indeed conveyed, from the idea of a distributed intelligent system as a collection of interacting software processes, to that of a network of physical agents that take information from the environment and act within the environment itself to change it. What will be meant by a distributed intelligent robot is the smart interconnection of heterogeneous units having different sensing, computation, and actuation abilities.

It is also foreseeable that different robots produced by different makers may be able to act in a near future within the same environment and capitalize on inter–robot cooperation, which feeds the image of a "society" of robots [10]. For this becoming real various important issues are still to be solved, concerning e.g. the definition of a "standard" cooperation language that is sufficiently rich to describe the behavior of each individual robotic agent as well as robot–robot interaction. Such a language should enable us to describe how e.g. two robots with different mechanical structures can jointly and efficiently manipulate an object. While standards are common practice in Information Technology, the larger variety and more complexity of these systems may be the causes that have prevented this in Robotics up to now. The need for *de facto* standards is more and more perceived at every application level [11]. With this aim, we provide a general model of a cooperating physical agent endowed with the ability to interact with other neighboring agents according to a set of event–based rules. The model is a hybrid system with a time–driven dynamics describing the behavior of the physical plant interconnected with an event–driven dynamics, in the form of an automaton, describing agent–agent cooperation.

Moreover, agent cooperation is advantageous in many applications as it enables desirable properties such as scalability, reconfigurability, and robustness (see e.g. [12], [13]). However, the fact that a distributed robot may be composed of many units, that are not secured in a protected environment with restricted accessibility permissions, makes it appealing for an attacker to compromise some of these units so as to degrade the system's QoS or to lead it to an unsafe condition. As an example, consider a group of cars with automated pilots that are supposed to move in a highway by following driving rules to avoid collisions. By tampering a pilot, it would be possible to generate a traffic jam or make a car crash into another. The whole system may be at risk if some of its units deviate from specification [14], which makes securing cooperative physical systems a major goal.

In this paper we address the misbehavior detection problem for a class of cooperative multi–robot systems, where the behavior of every agent depends on the presence or absence of other neighboring agents. As in a human society, misbehavior detection can not be achieved by supervision of a central authority only. It rather requires that every individual participate in monitoring its neighbors and share locally estimated "evidences" of their correctness, so that a globally accepted reputation of them can be established. Assessing an agent's reputation is a well–understood problem in Peer–To–Peer (P2P) systems and in Mobile Ad–hoc Networks (MANET). Uncooperative nodes not forwarding messages to their recipients, thus preventing an efficient routing service, can indeed be discouraged by e.g. the CONFIDANT protocol [15], [16].

To assess an agent's reputation by local observations is more complex in our context for various reasons. First, based on local sensing, an agent typically knows the set of another agent's neighbors only partially. In the highway example, some cars affecting the behavior of a target agent can be too far or hidden by other cars. The observed behavior of an agent is uncooperative if, and only if, it cannot be "explained" by the cooperative model for some inputs. This requires either full knowledge of the model's inputs, which is unlikely, or understanding if there exists a choice for the configurations of possible non–visible agents which explains that behavior. The latter case involves inverting the hybrid nonlinear model, which is effectively viable for specific systems only [17]–[19]. To cope with this, we abstract the inversion at the level of the cooperation automaton that has finite state and input domains.

Secondly, each agent's visibility and interaction topology are time–varying and unknown for a local monitoring agent. Therefore, available approaches to fault detection that are based on fixed and a–priori known event measurability [20]–[27] cannot be applied directly. Constructing a different observer for any possible combination of visibility and interaction topology is indeed inefficient, while we show how they can be encoded into a unique observer that is valid also when the event uncertainty is time–varying.

Misbehaving software agents in Computer Science, termed *intruders*, are discouraged in traditional settings by the implementation of an Intrusion Detection System (IDS) [28]. More recently, cryptography–based approaches, such as LARK [29], have been proposed to isolate malicious agents from communication in networked embedded systems. In this work, we focus on detecting agents that misbehave from the motion viewpoint and propose the architecture of a distributed IDS that is based on two components: a local monitoring process and a set–valued consensus protocol. The local monitor is able to reconstruct a map of free and occupied regions in the neighborhood of a target agent, while the consensus protocol allows all agents to reach a unique global view of the occupancy map and therefore on the reputation of a target agent. To this aim, available solutions based on linear consensus algorithms (see e.g. [30]–[32]) are inapplicable, as local monitors' outputs are continuous sets representing regions where the presence of other agents is required. The protocol is based on the assumption of a virtuous scenario, where the exchange of information between agents is correct, which can be guaranteed by the adoption of trusted software platforms [33]. No collusion exists between a robot executing an incorrect motion and another one trying to justify it. The problem of reaching consensus on information corrupted by intruders is a classical one [34] and is not investigated here. Obviously, detecting simultaneous motion and information misbehaviors is much more complex and is left for further studies. A preliminary study was presented in [35].

What makes our approach appealing are the facts the architecture of the IDS is general and the IDS can be automatically generated once the cooperation protocol is given. The paper is based on previous works by the authors [36]–[38], where only the fundamental ideas where presented, and is extended here with a complete formalization of the cooperative model and of the IDS. Two examples are also in depth considered to show the effectiveness of the proposed technique.

The paper is organized as follows. The general model of a cooperative physical agent is formalized as a distributed protocol in Section II. The architecture of a local monitor is described in Section III, while the set–valued consensus protocol is presented in Section IV. A formal description of the highway example is reported in the Appendix along with a proof of correctness of the local observer.

## II. A Model of Cooperation Protocols for Physical Agents

Consider a system composed of $n$ agents, $\mathcal{A}_1, \ldots, \mathcal{A}_n$, sharing a state-space, or *environment* $\mathcal{Q}$. By *cooperation protocol* we mean a formal description of the agents constitutive elements, i.e. their perceptions and actions, and of the rules used to interconnect these elements. More precisely, a *cooperation protocol* $\mathcal{P}$ consists in specifying for each agent $\mathcal{A}_i$ a nonuple $\mathcal{P}_i := \{f_i, \mathcal{V}_i, T_i, E_i, e_i, \Sigma_i, \delta_i, u_i\}$, where:

- $f_i : \mathcal{Q} \times \mathcal{U}_i \to T\mathcal{Q}$ is the *dynamics* map of $\mathcal{A}_i$.
  Here, $\mathcal{U}_i$ denotes the set of admissible input values and $T\mathcal{Q}$ the space tangent to $\mathcal{Q}$. The agent's state $q_i \in \mathcal{Q}$ evolves from its initial state $q_i^0$ according to the ODE
  $$\begin{cases} \dot{q}_i(t) = f_i(q_i(t), u_i(t)) \\ q_i(0) = q_i^0 \end{cases}, \; t \geq 0.$$

- $\mathcal{V}_i : \mathcal{Q}^n \to 2^{\mathcal{Q}}$ is the *visibility map* of $\mathcal{A}_i$, describing the region observed by its sensors;

- $T_i = \{\eta_{i,1}, \cdots, \eta_{i,\kappa_i}\}$ is a set of *topologies* on $\mathcal{Q}$, with $\eta_{i,j} : \mathcal{Q} \to 2^{\mathcal{Q}}$.
  Topologies are basic to define a few further concepts. We define an agent's *neighborhood* as $N(q_i) = \cup_{j=1}^{\kappa_i} \eta_{i,j}(q_i)$; a *neighbor set* as $M_i = \{\mathcal{A}_k \in \{\mathcal{A}_1 \cdots, \mathcal{A}_n\} \,|\, q_k \in N(q_i)\}$; a *neighbor configuration set* as $I_i = \{q_k \in \mathcal{Q} \,|\, \mathcal{A}_k \in M_i\}$. Finally, we define an agent's *encoder map* as $s_i : \mathcal{Q} \times \mathcal{Q}^{n_i} \to \mathbb{B}^{\kappa_i}$, where $n_i = \mathrm{card}(I_i)$ and $\mathbb{B} \stackrel{\text{def}}{=} \{0,1\}$. The $j$–th encoder component, $s_{i,j}$, is a logical–valued function returning 1 in the presence of an agent in the $j$–th topology $\eta_{i,j}(q_i)$, i.e.
  $$\begin{array}{rl} s_{i,j} : & \mathcal{Q} \times \mathcal{Q}^{n_i} \to \mathbb{B} \\ & (q_i, I_i) \mapsto \sum_{q_k \in I_i} \mathbf{1}_{\eta_{i,j}(q_i)}(q_k) \end{array},$$
  where $\sum$ represents the logical sum (*or*), and $\mathbf{1}_A(x)$ is the Indicator function of a set $A$;

- $E_i = \{e^{i,1}, \cdots, e^{i,\nu_i}\}$ is a finite *alphabet of events*;
- $e_i$ is an *event detector map*

$$\begin{aligned} e_i \; : \; & \mathbb{B}^{\kappa_i} \to 2^{E_i} \\ & s_i \mapsto \{e^{i,j} \in E_i \,|\, c_{i,j}(s_i) = 1\}\,, \end{aligned}$$

where each detector condition $c_{i,j}$ is a logical function of the form

$$\begin{aligned} c_{i,j} \; : \; & \mathbb{B}^{\kappa_i} \to \mathbb{B} \\ & s_i \mapsto \Pi_{k \in \gamma_{i,j}} s_{i,k} \, \Pi_{k \in \rho_{i,j}} \neg s_{i,k} \cdot \\ & \cdot \Pi_{k \in \mu_{i,j}} \mathbf{1}_{\lambda_{i,k}}(q_i) \, \Pi_{k \in \nu_{i,j}} \neg \mathbf{1}_{\lambda_{i,k}}(q_i)\,, \end{aligned} \qquad (1)$$

with $\lambda_{i,1}, \cdots, \lambda_{i,h_i}$ constants in $2^{\mathcal{Q}}$, $\gamma_{i,j} \cup \rho_{i,j} = \{1, \cdots, \kappa_i\}$ and $\gamma_{i,j} \cap \rho_{i,j} = \emptyset$, $\mu_{i,j} \cup \nu_{i,j} = \{1, \cdots, h_i\}$ and $\mu_{i,j} \cap \nu_{i,j} = \emptyset$, and $\Pi$ and $\neg$ the logical product (*and*) and negation (*not*), respectively;
- $\Sigma_i = \{\sigma^{i,1}, \ldots, \sigma^{i,p}\}$ is a finite *discrete state set*;
- $\delta_i \,:\, \Sigma_i \times 2^{E_i} \to \Sigma_i$ is a deterministic *automaton* describing how the agent's discrete state is updated, i.e.

$$\begin{cases} \sigma_i(t_{k+1}) = \delta_i(\sigma_i(t_k), e_i(t_{k+1}))\,, & t_k > 0 \\ \sigma_i(0) = \sigma_i^0 \end{cases},$$

where $\sigma_i^0 \in \Sigma_i$ is the initial discrete state, and $t_k$ is the $k$–th instant $t$ at which $e_i$ detects a new event [39];
- $u_i \,:\, \mathcal{Q} \times \Sigma_i \to \mathcal{U}_i$ is a control *decoder map*, describing which control value is applied at different states of the system, i.e.

$$u_i(t) = u_i(q_i(t), \sigma_i(t_k))\,.$$

According to this definition of cooperation protocol, the state $(q_i, \sigma_i) \in \mathcal{Q} \times \Sigma_i$ of a cooperative agent $\mathcal{A}_i$ following protocol $\mathcal{P}$ evolves as

$$\begin{aligned} \dot{q}_i(t) &= f_i(q_i(t), u_i(q_i(t), \sigma_i(t_k))) = f_i^*(q_i(t), \sigma_i(t_k))\,, \\ \sigma_i(t_{k+1}) &= \delta_i(\sigma_i(t_k), e_i(s_i(q_i(t), I_i(t)))) = \\ &= \delta_i^*(\sigma_i(t_k), q_i(t), I_i(t))\,, \end{aligned}$$

which can be written more compactly as

$$\begin{cases} (\dot{q}_i(t), \sigma_i(t_{k+1})) = \mathcal{H}_i(q_i(t), \sigma_i(t_k), I_i(t))\,, \\ (q_i(0), \sigma_i(0)) = (q_i^0, \sigma_i^0)\,, \end{cases} \qquad (2)$$

where $\mathcal{H}_i \,:\, \mathcal{Q} \times \Sigma_i \times \mathcal{Q}^{n_i} \to T_{\mathcal{Q}} \times \Sigma_i$ is the agent's *hybrid dynamic map* (cf. [40]). The *behavior* of each cooperative agent up to the current time $t$ is described by the solution $\phi_{\mathcal{H}_i}(q_i^0, \sigma_i^0, \tilde{I}_i(t))$ of the system in Eq. 2 subject to the input $\tilde{I}_i(t)$, intended as the history of its neighbor configuration set $I_i(\tau)$ for $\tau = 0, \cdots, t$.

In the following, we will assume that sensors are available on all agents so that each has complete knowledge of its own neighborhood, i.e. $\mathcal{V}_i(q_1, \cdots, q_n) \supseteq N(q_i)$.

### A. Examples: Robotic Warehouse and Automated Highway

Consider $n$ autonomous forklifts that are used to move products realized in a factory from carrier tapes to storage piles. Each forklift $\mathcal{A}_i$ is assigned a path that it must travel, when possible, at maximum speed $v_{max}$. To avoid collisions, it is supposed to give way to other forklifts approaching from a crossing path on its right and that are detected by an onboard, 360–degree camera with visibility range $R_i$ (Fig. 1–a).

The system's cooperation protocol $\mathcal{P}$ can be described as follows. The environment is $\mathcal{Q} = R^2 \times SO(2) \times R$. An agent state is $q_i = (x_i, y_i, \theta_i, v_i)$ and, based on its input $u_i = (a_i, \omega_i)$, is updated through the dynamic map

$$\begin{aligned} f_i \; : \; & \mathcal{Q} \times \mathcal{U}_i \to T_{\mathcal{Q}} \\ & (q_i, u_i) \mapsto (v_i \cos \theta_i,\, v_i \sin \theta_i,\, \omega_i,\, a_i)^T\,, \end{aligned}$$

The topology set is

$$\begin{aligned} \eta_{i,1} \; : \; & \mathcal{Q} \to 2^{\mathcal{Q}} \\ & q_i \mapsto \Big\{(x, y, \theta, v) \in \mathcal{Q} \,\big|\, (x - x_i)^2 + (y - y_i)^2 \leq d_i, \\ & \quad -\tfrac{\pi}{2} \leq \arctan\!\left(\tfrac{y - y_i}{x - x_i}\right) - \theta_i \leq \tfrac{\pi}{4} \Big\} \end{aligned}$$

where $d_i$ is a safety distance, and the corresponding encoder map is $s_i = s_{i,1}$ ($\kappa_i = 1$) with

$$\begin{aligned} s_{i,1} \; : \; & \mathcal{Q} \times \mathcal{Q}^{n_i} \to \mathbb{B} \\ & (q_i, I_i) \mapsto \textstyle\sum_{q_k \in I_i} \mathbf{1}_{\eta_{i,1}(q_i)}(q_k)\,. \end{aligned}$$

Thus, the agent's neighborhood is $N(q_i) = \eta_{i,1}(q_i)$. The event alphabet is $E_i = \{e^{i,1}, e^{i,2}\}$ and the detector map $e_i \in \mathbb{B} \to 2^{E_i}$, with $2^{E_i} = \{\emptyset, e^{i,1}, e^{i,2}, \{e^{i,1}, e^{i,2}\}\}$, is characterized by the event conditions $c_{i,1}, c_{i,2} \,:\, \mathbb{B} \to \mathbb{B}$, with $\gamma_{i,1} = \emptyset$, $\rho_{i,1} = \{1\}$, $\mu_{i,1} = \nu_{i,1} = \emptyset$, $\gamma_{i,2} = \{1\}$, $\rho_{i,2} = \mu_{i,1} = \nu_{i,1} = \emptyset$ ($\lambda_{i,j}$ need not be defined), i.e.,

$$c_{i,1} = \neg s_{i,1}\,, \quad c_{i,2} = s_{i,1}\,,$$

and thus

$$\begin{aligned} e_i \; : \; & \mathbb{B} \to 2^{E_i} \\ & 0 \mapsto \{e^{i,1}\}\,, \quad 1 \mapsto \{e^{i,2}\}\,. \end{aligned}$$

The finite set of discrete states is $\Sigma_i = \{\mathsf{ACC}, \mathsf{DEC}\}$ ($p = 2$) and the automaton's dynamics is

$$\begin{aligned} \delta_i \; : \; & \Sigma_i \times 2^{E_i} \to \Sigma_i \\ & (\mathsf{ACC}, e^{i,1}) \mapsto \mathsf{ACC}\,, \\ & (\mathsf{ACC}, e^{i,2}) \mapsto \mathsf{DEC}\,, \\ & (\mathsf{DEC}, e^{i,1}) \mapsto \mathsf{ACC}\,, \\ & (\mathsf{DEC}, e^{i,2}) \mapsto \mathsf{DEC}\,, \end{aligned}$$

with initial state $\sigma_i^0 = \mathsf{DEC}$. The decoder map is

$$\begin{aligned} u_i \; : \; & \mathcal{Q} \times \Sigma_i \to \mathcal{U}_i \\ & (q_i, \mathsf{ACC}) \mapsto (-\mu(v_i - v_{max}), 0)^T\,, \\ & (q_i, \mathsf{DEC}) \mapsto (-\mu\, v_i, 0)^T\,, \end{aligned}$$

where $\mu$ is a positive constant, which implies that the configuration $q_i$ evolves according to the controlled dynamic map

$$\begin{aligned} f_i^* \; : \; & \mathcal{Q} \times \Sigma_i \to T_{\mathcal{Q}} \\ & (q_i, \mathsf{ACC}) \mapsto (v_i \cos \theta_i,\, v_i \sin \theta_i,\, 0,\, -\mu(v_i - v_{max}))\,, \\ & (q_i, \mathsf{DEC}) \mapsto (v_i \cos \theta_i,\, v_i \sin \theta_i,\, 0,\, -\mu\, v_i)\,. \end{aligned}$$

The solution $q_i(t) = \phi_{f_i^*}(q_i(t_k), \sigma_i(t_k))$, for $t \geq t_k$, of the controlled dynamics is

$$\begin{cases} x_i(t) &= x_i(t_k) + \Delta(\sigma_i(t_k), t) \cos(\theta_i(0))\,, \\ y_i(t) &= y_i(t_k) + \Delta(\sigma_i(t_k), t) \sin(\theta_i(0))\,, \\ \theta_i(t) &= \theta_i(0)\,, \\ v_i(t) &= V(\sigma_i(t_k), t) + v_i(t_k)\, e^{-\mu(t - t_k)}\,, \end{cases} \qquad (3)$$

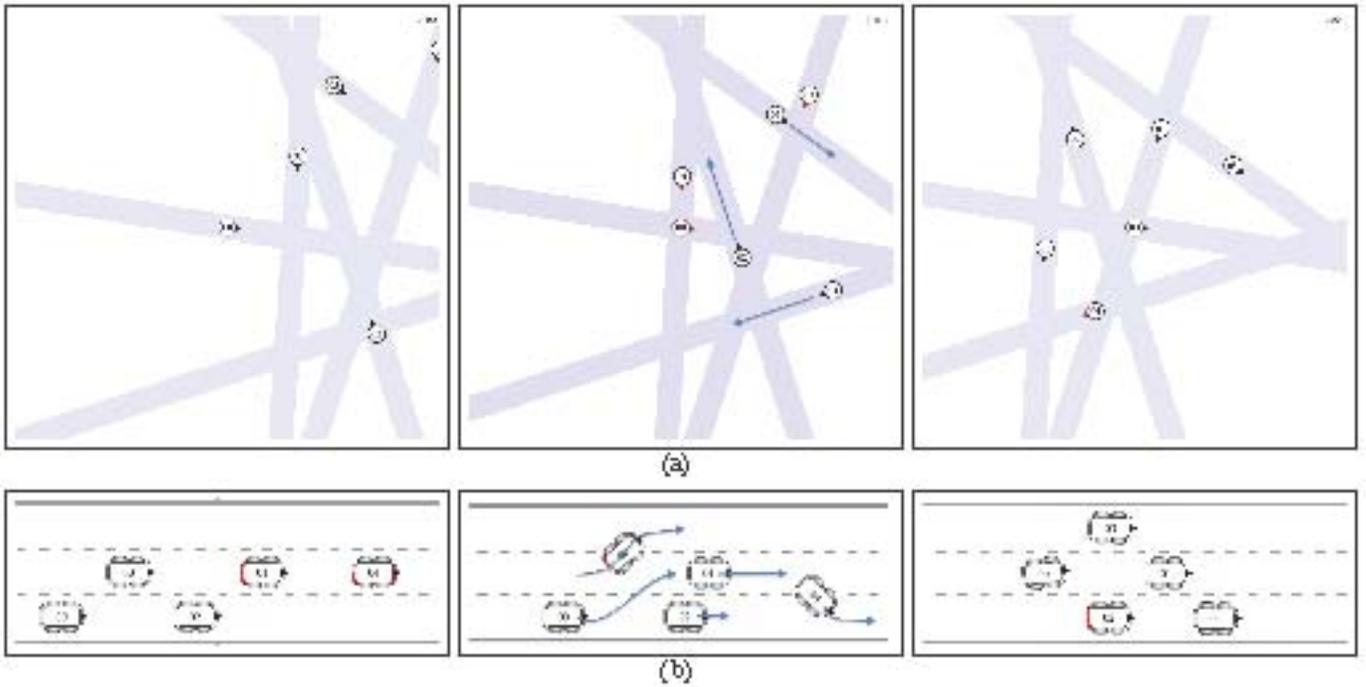

Figure 1. Examples of cooperative systems following a cooperation protocol $\mathcal{P}$: (a) Forklifts in a warehouse avoiding robot-robot and robot-human collisions by giving way on their own's right; (b) Cars in a highway following the common driving rules to avoid collisions.

with

$$\Delta(\text{ACC}, t) = z_{max}(t - t_k) + \frac{z_i(t_k) - z_{max}}{\mu}\left(1 - e^{-\mu(t - t_k)}\right),$$
$$\Delta(\text{DEC}, t) = \frac{z_i(t_k)}{\mu}\left(1 - e^{-\mu(t - t_k)}\right),$$
$$V(\text{ACC}, t) = z_{max}\left(1 - e^{-\mu(t - t_k)}\right), \quad V(\text{DEC}, t) = 0.$$

Finally, the visibility map is

$$\mathcal{V}_i : \mathcal{Q}^n \to 2^{\mathcal{Q}}$$
$$(q_1, \ldots, q_n) \mapsto \{q \in \mathcal{Q} \mid (x - x_i)^2 + (y - y_i)^2 \le R_i\},$$

with $R_i > d_i$.

A more elaborate example of cooperative system where interaction among agents can be formalized as an instance of the cooperation protocol $\mathcal{P}$ is represented by a group of $n$ cars in a highway (Fig. 1-b). Cars have their own dynamics $f_i$ and local controllers $z_i$, but their pilots are supposed to follow a set of traffic rules to avoid collisions. Each car is supposed to decide a suitable maneuver, i.e., accelerate (FAST) or decelerate (SLOW), change to the next left lane (LEFT) or to the right one (RIGHT), based on the presence or absence of other cars in its neighborhood. E.g., the presence of a slower car in the front, and a free lane on the left requires the execution of an overtake that is a change from a FAST to a LEFT maneuver. The rules require the introduction of a topology $\eta_{i,1}(q_i)$ representing a region in the immediate front of an agent $\mathcal{A}_i$, a topology $\eta_{i,2}(q_i)$ for a region on its left, a topology $\eta_{i,3}(q_i)$ for a region on its right, and a topology $\eta_{i,4}(q_i)$ for a region on its back. A complete description of the system's formalization is reported in Appendix A. Note that these rules apply to very large systems with an unbounded number of vehicles, yet they only require that every car verifies the existence and/or absence of $N$ other cars in its vicinity, where $N$ is a small number depending only on the geometry of the lanes and of the vehicles.

## III. OBSERVER-BASED INTRUSION DETECTION PROTOCOL - LOCAL MONITOR

Consider an agent $\mathcal{A}_h$ trying to learn whether another agent $\mathcal{A}_i$ is cooperative or not. The basic difficulty in classifying the behavior of $\mathcal{A}_i$ arises from the incomplete view of its neighborhood, as it generally holds that $N(q_i) \not\subseteq \mathcal{V}_h(q_1, \ldots, q_n)$. Indeed, consider again a scenario where a local observer on-board car 03 is trying to learn whether the car 00 is cooperative or not. Not having full access to the information available for the car 00 itself, it is difficult for the observer to decide whether the pilot is correctly driving or if it is simulating the presence of another car, which would be hidden to the observer's view or outside of its range of visibility. In these cases, a simple approach based on comparison of the actual trajectory of $\mathcal{A}_i$, $\tilde{q}_i(t)$, with the solution of the cooperative model (Eq. 2) cannot be employed, because the input $\tilde{I}_i(t)$ is only partially known to $\mathcal{A}_h$. To cope with this issue, our approach uses a local dynamic process, called *monitor*, which is composed of two components: a *hybrid observer* of the state $(q_i, \sigma_i)$ of $\mathcal{A}_i$, and an *occupancy estimator* in the $\mathcal{A}_i$ agent's neighborhood $N(q_i)$.

### A. The Hybrid Observer

The hybrid observer execution consists of two distinct phases, *predict* and *update*, which are run every observation period $T_k = [t_k, t_{k+1}]$. During the predict phase, a priori estimates $(\hat{q}_i(t|t_k), \hat{\sigma}_i(t_k|t_k))$ of the trajectories that $\mathcal{A}_i$ can execute during the current period are computed, based on previous estimates and the locally visible neighbor configuration set $\hat{I}_i^h = I_i \cap \mathcal{V}_h(q_1, \ldots, q_n)$. This can be achieved by introducing the following objects, which are defined for every agent running a cooperation protocol $\mathcal{P}$:

- A *topology check* map $v_i : \mathcal{Q} \times 2^{\mathcal{Q}} \to \mathbb{B}^{\kappa_i}$, returning a binary vector whose $j$-th entry is 1 iff the $j$-th topology is entirely visible from agent $\mathcal{A}_h$, i.e.,

$$v_{i,j} : \mathcal{Q} \times 2^{\mathcal{Q}} \to \mathbb{B}$$
$$(q_i, V_h) \mapsto \begin{cases} 1 & \text{if } \eta_{i,j}(q_i) \subseteq V_h \\ 0 & \text{otherwise} \end{cases} ;$$

- A *restricted encoder* map $\tilde{s}_i : \mathcal{Q} \times \mathcal{Q}^{\hat{n}_i} \to \mathbb{B}^{\kappa_i}$, where $\hat{n}_i = \text{card}(I_i^h)$, whose $j$-th component returns a lower approximation of $s_{i,j}$ based on the locally visible neighbor configuration set $I_i^h$ as

$$\tilde{s}_{i,j} : \mathcal{Q} \times \mathcal{Q}^{\hat{n}_i} \to \mathbb{B}$$
$$(q_i, I_i^h) \to \sum_{q_k \in I_i^h} \mathbf{1}_{\eta_{i,j}(q_i)}(q_k)$$

(note that $\tilde{s}_{i,j}(q_i, I_i^h) \leq s_{i,j}(q_i, I_i)$ since $I_i^h \subseteq I_i$);

- An *event estimator* map $\tilde{e}_i : \mathbb{B}^{\kappa_i} \times \mathbb{B}^{\kappa_i} \to 2^{E_i}$, $\tilde{e}_i(\hat{s}_i, v_i)$ that provides an upper approximation of the event detector output $e_i(s_i)$;

- A *nondeterministic automaton*

$$\tilde{\delta}_i : 2^{\Sigma_i} \times 2^{E_i} \to 2^{\Sigma_i}$$
$$(\hat{\sigma}_i, \hat{e}_i) \to \{\bar{\sigma} \in \Sigma_i \,|\, \exists \sigma \in \Sigma_i, \sigma \subseteq \hat{\sigma}_i \,|$$
$$|\, \delta_i(\sigma, \hat{e}_i) = \bar{\sigma}\},$$

describing how the agent's discrete state estimate $\hat{\sigma}_i \in 2^{\Sigma_i}$ is updated as

$$\begin{cases} \hat{\sigma}_i(t_{k+1}) = \tilde{\delta}_i(\hat{\sigma}_i(t_k), \hat{e}_i(t_{k+1})) \\ \hat{\sigma}_i(0) = \hat{\sigma}_i^0 \end{cases},$$

where $\hat{\sigma}_i^0$ is the initial estimate;

- A *controlled dynamics* map

$$\tilde{f}_i^* : 2^{\mathcal{Q}} \times 2^{E_i} \to 2^{T\mathcal{Q}}$$
$$(\hat{q}_i, \hat{\sigma}_i) \mapsto \{\dot{q} \in T_{\mathcal{Q}} \,|\, \exists \bar{q} \in \mathcal{Q}, \bar{q} \subseteq \hat{q}_i,$$
$$\exists \bar{\sigma} \in \Sigma_i, \bar{\sigma} \subseteq \hat{\sigma}_i \,|\, f_i^*(\bar{q}, \bar{\sigma}) = \dot{q}\},$$

describing how the agent's estimated configuration $\hat{q}_i \in 2^{\mathcal{Q}}$ is updated

$$\begin{cases} \dot{\hat{q}}_i(t) = \tilde{f}_i^*(\hat{q}_i(t), \hat{\sigma}_i(t_k)) \\ \hat{q}_i(0) = \hat{q}_i^0 \end{cases},$$

with $\hat{q}_i^0$ the initial estimate.

Clearly, the crucial step to obtain a valid observer is the event estimator map, which should provide as tight an approximation as possible. To this purpose, the following result (whose proof is reported in Appendix B) is instrumental.

*Theorem 1:* The smallest event estimator compatible with an available topology information $v_i$ and an encoder map $\hat{s}_i$ is given by

$$\tilde{e}_i : \mathbb{B}^{\kappa_i} \to 2^{E_i}$$
$$\hat{s}_i \mapsto \{e^{i,j} \in E_i \,|\, \tilde{c}_{i,j}(\hat{s}_i, v_i) = 1\},$$

with

$$\tilde{c}_{i,j} : \mathbb{B}^{\kappa_i} \times \mathbb{B}^{\kappa_i} \to \mathbb{B}$$
$$(\hat{s}_i, v_i) \to \Pi_{k \in \gamma_{i,j}} (\hat{s}_{i,k} v_{i,k} + \neg v_{i,k}) \cdot$$
$$\cdot \Pi_{k \in \rho_{i,j}} \neg \hat{s}_{i,k} \cdot \qquad (4)$$
$$\cdot \Pi_{k \in \mu_{i,j}} \lambda_{i,k} \Pi_{k \in \tau_{i,j}} \neg \lambda_{i,k} ;$$

*1) Predict Phase:* A nondeterministic hybrid model to predict every behavior that $\mathcal{A}_i$ can execute, compatible with information locally available to $\mathcal{A}_h$, is given by

$$\dot{\hat{q}}_i(t) = \tilde{f}_i^*(\hat{q}_i(t), \hat{\sigma}_i(t_k)),$$
$$\hat{\sigma}_i(t_{k+1}) = \tilde{\delta}_i(\hat{\sigma}_i(t_k), \tilde{e}_i(\hat{s}_i(t_{k+1}), \hat{v}_i(t_{k+1}))) =$$
$$= \tilde{\delta}_i^*(\hat{\sigma}_i(t_k), \bar{q}_i(t), I_i^h(t)),$$

where $\hat{v}_i(t_k) = v_i(\bar{q}_i(t_k), \mathcal{V}_h(q_1(t), \cdots, q_n(t)))$ and $\hat{s}_i(t_k) = \tilde{s}_i(\bar{q}_i(t_k), I_i^h(t_k))$, which can be rewritten more compactly as

$$\begin{cases} (\dot{\hat{q}}_i(t), \hat{\sigma}_i(t_{k+1})) = \tilde{\mathcal{H}}_i(\hat{q}_i(t), \hat{\sigma}_i(t_k), I_i^h(t)), \\ (\hat{q}_i(0), \hat{\sigma}_i(0)) = (\hat{q}_i^0, \hat{\sigma}_i^0), \end{cases}$$

where $\tilde{\mathcal{H}}_i : 2^{\mathcal{Q}} \times 2^{\Sigma_i} \times \mathcal{Q}^{\hat{n}_i} \to 2^{T\mathcal{Q}} \times 2^{\Sigma_i}$. The a priori estimate of $\mathcal{A}_i$'s state is initialized with the values $\hat{q}_i(t_0|t_0) = \bar{q}_i(t_0)$ and $\hat{\sigma}_i(t_{-1}|t_0) = \Sigma_i$ (the whole discrete state set, i.e. the most conservative hypothesis). The propagation law is

$$\hat{q}_i(t|t_k) = \phi_{\tilde{f}_i^*}(\hat{q}_i(t_k|t_k), \hat{\sigma}_i(t_k|t_k), \tilde{I}_i^h(t)),$$
$$\hat{\sigma}_i(t_k|t_k) = \tilde{\delta}_i(\hat{\sigma}_i(t_{k-1}|t_k), \tilde{e}_i(\hat{s}_i(t_k|t_k), \hat{v}_i(t_k))),$$

for $t \geq t_k$, where $\hat{s}_i(t_k|t_k) = \tilde{s}_i(\bar{q}_i(t_k), I_{i,j}^h(t_k))$ is the a priori estimate of the topology activation, and $\tilde{I}_i^h(t)$ is the history of $I_i^h$ from $t_k$ to $t$. To avoid explicit model inversion during the following update phase (which is impractical for general systems), the set describing the "relationship" between predicted configuration trajectories and discrete states is computed as

$$L(t_k) = \hat{q}_i(t|t_k) \bowtie_{\tilde{f}_i^*} \hat{\sigma}_i(t_k|t_k),$$

where we introduce the operator $\bowtie_{\tilde{f}_i^*}$ defined as

$$\bowtie_{\tilde{f}_i^*} : 2^{\mathcal{Q}} \times 2^{\Sigma_i} \to 2^{\mathcal{Q} \times \Sigma_i}$$
$$(\hat{q}_i(t), \hat{\sigma}_i) \mapsto \{(q(t), \sigma) \,|\, q(t) \subseteq \phi_{\tilde{f}_i^*}(\hat{q}_i(t_k), \sigma)\}.$$

*2) Update Phase:* During the update phase, the a priori prediction is combined with current observation to refine the state estimate into an a posteriori state estimate. The update step, which is run at the end of the observation period, is given by

$$\hat{q}_i(t_{k+1}|t_{k+1}) = \bar{q}_i(t_{k+1}),$$
$$\hat{\sigma}_i(t_k|t_{k+1}) = \pi_{\Sigma_i}(L(t_k) \bowtie_\epsilon \bar{q}_i(t_k)),$$

where $\pi_A$ is the projector over the set $A$, and $\bowtie_\epsilon$ is defined as

$$\bowtie_\epsilon : 2^{\mathcal{Q} \times \Sigma_i} \times \mathcal{Q} \to 2^{\mathcal{Q} \times \Sigma_i}$$
$$(L, \bar{q}_i(t)) \mapsto \{(q(t), \sigma) \in L \,|\, ||q(t) - \bar{q}_i(t)|| \leq \epsilon\}.$$

Here, $\epsilon$ is a tolerance parameter, to be set depending on the accuracy of available sensors and nominal models. Finally, an a posteriori estimate of the encoder map outputs can be computed as

$$\hat{s}_i(t_k|t_{k+1}) = \{s^* \in \mathbb{B}^{\kappa_i} \,|\, s^* \geq \hat{s}_i(t_k|t_k),$$
$$e_i(s^*) \subseteq \hat{\sigma}_i(t_k|t_k) \bowtie_{\tilde{\delta}_i} \hat{\sigma}_i(t_k|t_{k+1}),$$

where

$$\bowtie_{\tilde{\delta}_i} : 2^{\Sigma_i} \times 2^{\Sigma_i} \to 2^{E_i}$$
$$(\hat{\sigma}_i, \hat{\sigma}_i^+) \mapsto \{\hat{e}_i \in E_i \,|\, \exists \bar{\sigma}, \bar{\sigma}^+ \in \Sigma_i,$$
$$\bar{\sigma} \subseteq \hat{\sigma}_i, \bar{\sigma}^+ \subseteq \hat{\sigma}_i^+, \bar{\sigma}^+ \subseteq \tilde{\delta}_i(\bar{\sigma}, e)\}.$$

*B. Occupancy Estimator*

Based on the visibility of an observer onboard $\mathcal{A}_h$, each component $s_{i,k}$ of the encoder map $s_i$ can be decomposed into the sum of the same component of the restricted encoder map $\tilde{s}_{i,k}$ and an unknown binary variable activating on the presence of agents $\mathcal{A}_k$, $k \neq i, h$ in the portion of the $k$–th topology $\eta_{i,k}$ of $\mathcal{A}_i$, that is out of the observer visibility region $\mathcal{V}_h$, i.e.,

$$s_{i,k}(q_i, I_i) = \tilde{s}_{i,k}(q_i, I_i^h) + p_{i,k}(q_i, I_i),$$

where

$$p_{i,k}(q_i, I_i) = \sum_{q_j \in I_i \setminus V_h} \mathbf{1}_{\eta_{i,k}(q_i)}(q_j).$$

An estimate of the logical vector $p_i = (p_{i,1}, \cdots, p_{i,\kappa_i})^T$ can be obtained, by combining the a priori and the a posteriori estimates of the encoder map, as described in the estimator map

$$\tilde{p}_i : \mathbb{B}^{\kappa_i} \times 2^{\mathbb{B}^{\kappa_i}} \to 2^{\mathbb{B}^{\kappa_i}},$$

whose $k$–th component is

$$\tilde{p}_{i,k} : \mathbb{B} \times 2^{\mathbb{B}} \to 2^{\mathbb{B}}$$
$$(0,0) \mapsto 0,\ (0,1) \mapsto 1,\ (0,\{0,1\}),\ (1,1) \mapsto \{0,1\}.$$

Note that $\tilde{p}_{i,k}$ is undefined for the inputs $(1,0)$, $(1,\{0,1\})$ as the a posteriori estimate of $s_{i,k}$ is obviously greater than the a priori ones.

Let us denote with $\hat{p}_i = \tilde{p}_i(\hat{s}_i(t_k|t_k), \hat{s}_i(t_k|t_{k+1}))$, and with $\bar{I}_i^h = \{\hat{q}_k \in 2^{\mathcal{Q}} \,|\, \exists\, q_k \in I_i^h \,|\, \hat{q}_k = B_\epsilon(q_k)\}$, where $B_\epsilon(q_k)$ is a ball with radius $\epsilon$ and center at $q_k$, an over–approximation of the monitor's measures taking into account of its sensor inaccuracy. An occupancy estimate of $\mathcal{A}_i$'s neighborhood, i.e. an estimate of its neighborhood configuration set $I_i$, can be obtained by using the map

$$\hat{I}_i^h : 2^{\mathcal{Q}} \times 2^{\mathcal{Q}} \times 2^{\mathbb{B}^{\kappa_i}} \to \mathcal{P}(\mathcal{Q})$$
$$(\bar{I}_i^h, V_h, \hat{p}_i) \mapsto$$
$$\{(\hat{\eta}_{i,j}(\bar{I}_i^h, V_h, p_{i,1}), \cdots, \hat{\eta}_{i,\kappa_i}(\bar{I}_i^h, V_h, p_{i,\kappa_i})$$
$$|\, p = (p_{i,1}, \cdots, p_{i,\kappa_i}) \in \mathbb{B}^{\kappa_i},\ p \subseteq \hat{p}_i\},$$

where $\mathcal{P}(\mathcal{Q}) = 2^{2^{\mathcal{Q}}}$, and the occupancy in the $k$–th region is

$$\hat{\eta}_{i,k} : 2^{\mathcal{Q}} \times 2^{\mathcal{Q}} \times \mathbb{B} \to 2^{\mathcal{Q}}$$
$$(\bar{I}_i^h, V_h, 0) \mapsto \bar{I}_i^h \cap \eta_{i,k}(q_i),$$
$$(\bar{I}_i^h, V_h, 1) \mapsto (\bar{I}_i^h \cap \eta_{i,k}(q_i)) \cup (\eta_{i,k}(q_i) \setminus V_h).$$

## IV. Set–Valued Consensus Protocol for Monitor Agreement

Consider $m_i$ robotic agents, $\mathcal{A}_{i_1}, \cdots, \mathcal{A}_{i_{m_i}}$, trying to *consent* on whether a common neighboring agent $\mathcal{A}_i$ is cooperative or not. This section shows how they can agree upon a global "view" of the occupancy map of $\mathcal{A}_i$'s neighborhood, $N(q_i)$, by allowing every agent $\mathcal{A}_h$ to share local estimates, $\hat{\eta}_{i,j}^h$, for all $j$, of $\mathcal{A}_i$'s topology set with neighbors of a communication network $\mathcal{G}$.

Available solutions for network agreement typically consist of consensus protocols where agents exchange messages containing real scalars or vectors, that are combined together via a simple rule described as a linear dynamic system of the form $x(k+1) = A x(k)$, where $x \in \mathbb{R}^{m_i}$ is the vector of every agent's local estimate and $A$ is a suitable real square matrix [41], [42]. However, as described above, outputs from local monitors are continuous sets that cannot be merged — at least trivially — by means of a linear combination rule. There are indeed network agreement problems which raise the question of how to consent on more complex data structures (see e.g. [43] and our current problem).

With this respect, consider the problem of how to design a consensus protocol within the following generalized framework. Let consider the configuration space $\mathcal{Q}$ as the consensus domain set. Let $X_h \in 2^{\mathcal{Q}}$ be the "consensus state" of an agent $\mathcal{A}_h$ representing its local estimate of the quantity over which the agreement is sought. Let also $F : 2^{\mathcal{Q}} \times 2^{\mathcal{Q}} \to 2^{\mathcal{Q}}$ be a *merging function* that, given any two states $X_h, X_k$, produces another state $F(X_h, X_k)$. It is straightforward to introduce the composed function

$$F^{(l)} : 2^{\mathcal{Q}^l} \to 2^{\mathcal{Q}}$$
$$(X_1, \cdots, X_l) \mapsto F(\cdots F(X_1, X_2) \cdots, X_l).$$

Consider also the

$$X^* = F^{(m_i)}(U_1, \cdots, U_{m_i}), \quad (5)$$

where $U_i = X_i(0)$, that can be interpreted as the decision of a hypothetical centralized entity that knows every agent's initial estimate. Under the hypotheses that $F$ is *commutative*, i.e., $F(X_1, X_2) = F(X_2, X_1)$ for all $X_1, X_2$, and *associative*, i.e., $F(X_1, F(X_2, X_3)) = F(F(X_1, X_2), X_3)$ for all $X_1, X_2, X_3$, $X^*$ is well–defined since it is independent of the order by which the estimates are processed.

Before going to the main result, let us recall that a communication network $\mathcal{G}$ can be described by a graph $G(V_G, E_G)$, where $V_G$ is a set of nodes representing the agents and $E_G$ is a set of edges connecting agents that are within communication range. We assume that the graph is *undirected*, i.e., if $\mathcal{A}_h$ can send a message to $\mathcal{A}_k$, then also the reverse holds. Let $CV_h(p) \stackrel{\text{def}}{=} \{j \in V_G \mid \text{dist}(i,j) \leq p\}$ be the set of the agents that can send a message to $\mathcal{A}_h$ by passing through at most $p$ other agents, and let $\text{dist}(h,k)$ be the geodesic distance from $\mathcal{A}_h$ and $\mathcal{A}_k$, i.e. the length of the shortest communication path between $\mathcal{A}_h$ and $\mathcal{A}_k$ (note that $\text{dist}(h,h) = 0$, $\forall h \in V_G$). Let us also recall the notion of graph diameter as the maximum distance between any two nodes in a graph, that is $\text{diam}(G) = \max_{i,j \in V_G} \text{dist}(i,j)$. Finally, recall that $F$ is said to be *idempotent* if $F(X_1, X_1) = X_1$ for all $X_1$.

We are now ready to prove the following result:

*Theorem 2 (Set–Valued Consensus Protocol):* A collection of $m_i$ agents running a consensus protocol described by the dynamic system

$$\begin{cases} X_h(k+1) = F^{(p_h(1))}(X_{h,1}(k), \cdots, X_{h,p_h(1)}(k)), \\ X_h(0) = U_h, \end{cases}$$
(6)

where $p_h(k) = \text{card}(CV_h(k))$, for all agents $h$, converges to the centralized consensus state in at most $\tilde{n} = \text{diam}(G)$ rounds, i.e.,

$$X(\tilde{n}) = \mathbf{1}_{m_i} X^*,$$

if $F$ is commutative, associative, and idempotent, and if the communication graph $G$ is connected.

*Proof:* Let us first prove that the consensus step of an agent $\mathcal{A}_h$ after $k$ consensus step is
$$X_h(k) = F^{(p_h(k))}(X_{h,1}(0), \cdots, X_{h,p_h(k)}(0)).$$
The property is trivially satisfied after one consensus step:
$$X_h(1) = F^{(p_h(1))}(X_{h,1}(0), \cdots, X_{h,p_h(1)}(0)).$$
We want to prove it by induction, i.e. by assuming that it holds after $k$ steps, we need to prove its validity also after $k+1$ steps. Indeed we have:
$$X_h(k+1) = F^{(p_h(1))}(J_1(k), \cdots, J_{p_h(1)}(k)),$$
where is $J_i(k) = F^{(p_i(k))}(X_{i,1}(0), \cdots, X_{i,p_i(k)}(0))$ by the inductive hypothesis. Moreover, note that the order by which every estimate is processed is irrelevant, by the associativity and commutativity properties of $F$, and that multiple occurrence of the same estimate $X_{i,j}(0)$ can be simplified by its idempotency. The last equation involves the set of all agents $l \in CV_j(k)$ of all agents $j \in CV_h(1)$, whose union gives by definition the set of agents that can send a message to $\mathcal{A}_h$ via a communication path of at most $k+1$ other agents, i.e. $CV_h(k+1)$, which proves the property.

To prove the theorem, it is sufficient to note that, for all $k \geq \tilde{n}$, $CV_h(k) = V_G$ and hence $p_h(k) = m_i$, as the communication graph $G$ is connected. Therefore, we have
$$\begin{aligned} X_h(k) &= F^{(p_h(\tilde{n}))}(X_{h,1}(0), \cdots, X_{h,p_h(\tilde{n})}(0)) = \\ &= F^{(m_i)}(U_1 \cdots, U_{m_i}) = X^*, \end{aligned}$$
for all $h$ and all $k \geq \tilde{n}$, which concludes the proof. ∎

Consider a hypothetical *centralized observer* $\mathcal{O}$ that is able to receive the neighbor configuration sets from all the $m_i$ agents, $\mathcal{A}_{i_1}, \cdots, \mathcal{A}_{i_{m_i}}$, that are monitoring $\mathcal{A}_i$. Let
$$I_i^c = \hat{I}_i^{(i_1)} \cup \hat{I}_i^{(i_2)} \cup \cdots \cup \hat{I}_i^{(i_{m_i})}$$
be the centralized estimated neighbor configuration set. Then, the main implication of Theorem 2 to our intrusion detection problem is stated in the following:

*Corollary 1:* If the misbehavior of an agent $\mathcal{A}_i$ can be detected by the centralized observer $\mathcal{O}$, the same result can be achieved in a fully distributed way by making all local monitors reach an agreement on the estimated occupancy maps for $\mathcal{A}_i$'s neighborhood.

In particular, monitors can start with initial estimates given by the locally estimated occupancy maps, i.e.,
$$U_h = \hat{I}_i^h(t_k|t_{k+1})$$
and they can use the following merging function
$$\begin{aligned} \cap^* \;:\; & 2^\mathcal{Q} \times 2^\mathcal{Q} \to 2^\mathcal{Q} \\ & (X_1, X_2) \mapsto \{x \mid \exists x_1 \in X_1 \setminus \emptyset, x_2 \in X_2 \setminus \emptyset \mid \\ & \qquad x = x_1 \cap x_2\}. \end{aligned}$$
where $\cap$ is the set–theoretic intersection, which satisfies the theorem's hypotheses.

## V. Examples

### A. The Warehouse Example (Cont'd) - Monitor Construction

Consider a forklift $\mathcal{A}_h$ trying to learn whether another forklift $\mathcal{A}_i$ is cooperative or not. The corresponding monitor is constructed as follows. The topology visibility map is $v_i = (v_{i,1})$, with
$$\begin{aligned} v_{i,1} \;:\; & \mathcal{Q} \times 2^\mathcal{Q} \to \mathbb{B} \\ & (q_i, V_h) \mapsto \begin{cases} 1 & \text{if } \eta_{i,1}(q_i) \subseteq V_h, \\ 0 & \text{otherwise}, \end{cases} \end{aligned}$$
and the encoder map is $\tilde{s}_i = \tilde{s}_{i,1}$ with
$$\begin{aligned} \tilde{s}_{i,1} \;:\; & \mathcal{Q} \times \mathcal{Q}^{\hat{n}_i} \to \mathbb{B} \\ & (q_i, I_i^h) \to \sum_{q_k \in I_i^h} \mathbf{1}_{\eta_{i,1}(q_i)}(q_k). \end{aligned}$$
The detection conditions are determined by the values of $\gamma_{i,j}$, $\rho_{i,j}$, $\mu_{i,j}$, and $\nu_{i,j}$, for $j = 1, 2$:
$$\begin{aligned} \tilde{c}_i \;:\; & \mathbb{B} \times \mathbb{B} \to \mathbb{B}^2 \\ & (\hat{s}_i, v_i) \mapsto \begin{pmatrix} \neg \hat{s}_{i,1} \\ \hat{s}_{i,1} v_{i,1} + \neg v_{i,1} \end{pmatrix}. \end{aligned}$$
The automaton's initial state is $\hat{\sigma}_i^0 = \{\text{ACC}, \text{DEC}\}$ and its nondeterministic dynamics is
$$\begin{aligned} \tilde{\delta}_i \;:\; & 2^{\Sigma_i} \times 2^{E_i} \to 2^{\Sigma_i} \\ & \begin{array}{l} (\text{ACC}, e^{i,1}), (\text{DEC}, e^{i,1}), \\ (\{\text{ACC}, \text{DEC}\}, e^{i,1}) \end{array} \mapsto \text{ACC}, \\ & \begin{array}{l} (\text{ACC}, e^{i,2}), (\text{DEC}, e^{i,2}), \\ (\{\text{ACC}, \text{DEC}\}, e^{i,2}) \end{array} \mapsto \text{DEC}, \\ & \begin{array}{l} (\text{ACC}, \{e^{i,1}, e^{i,2}\}), \\ (\text{DEC}, \{e^{i,1}, e^{i,2}\}), \\ (\{\text{ACC}, \text{DEC}\}, \{e^{i,1}, e^{i,2}\}) \end{array} \mapsto \{\text{ACC}, \text{DEC}\}, \end{aligned}$$
and the controlled dynamic map is
$$\begin{aligned} \tilde{f}_i^* \;:\; & 2^\mathcal{Q} \times 2^{\Sigma_i} \to 2^{T\mathcal{Q}} \\ & (\hat{q}_i, \text{ACC}) \mapsto \left( \hat{v}_i \cos^* \hat{\theta}_i, \hat{v}_i \sin^* \hat{\theta}_i, 0, -\mu(\hat{v}_i - v_{max}) \right) \\ & (\hat{q}_i, \text{DEC}) \mapsto \left( \hat{v}_i \cos^* \hat{\theta}_i, \hat{v}_i \sin^* \hat{\theta}_i, 0, -\mu \hat{v}_i \right), \\ & (\hat{q}_i, \{\text{ACC}, \text{DEC}\}) \mapsto \begin{pmatrix} \hat{v}_i \cos^* \hat{\theta}_i \\ \hat{v}_i \sin^* \hat{\theta}_i \\ 0 \\ \{-\mu(\hat{v}_i - v_{max}), -\mu \hat{v}_i\} \end{pmatrix}, \end{aligned}$$
where $\cos^*$ and $\sin^*$ are
$$\begin{aligned} \cos^* \;:\; & 2^\mathbb{R} \to 2^\mathbb{R} \\ & \hat{\alpha} \mapsto \{\alpha \in \mathbb{R} \mid \exists \bar{\alpha} \in \mathbb{R}, \bar{\alpha} \subseteq \hat{\alpha} \mid \cos(\bar{\alpha}) = \alpha\}, \\ \sin^* \;:\; & 2^\mathbb{R} \to 2^\mathbb{R} \\ & \hat{\alpha} \mapsto \{\alpha \in \mathbb{R} \mid \exists \bar{\alpha} \in \mathbb{R}, \bar{\alpha} \subseteq \hat{\alpha} \mid \sin(\bar{\alpha}) = \alpha\}. \end{aligned}$$

### B. The Warehouse Example (Cont'd) - Corrupted Encoder

Consider an attack undertaken by a misbehaving forklift $\mathcal{A}_i$ whose neighborhood $N(q_i)$ is free of other forklifts, i.e. $I_i(t) = \emptyset$ for all $t$, while the agent simulates the existence of a forklift $\mathcal{A}_j$ s.t. $q_j(t) \in \eta_{i,1}(q_i)$ for $t \geq 2T$. Hence, the neighbor configuration set that $\mathcal{A}_i$ pretends to be subject to is
$$I_i(t) = \begin{cases} \emptyset & t < 2T \\ q_j(t) & t \geq 2T \end{cases}.$$

The agent's encoder map $s_i(t) = s_{i,1}(t)$ correspondingly takes the values $s_{i,1}(t) = 0$, for $t \in [0, 2T)$ and $s_{i,1}(t) = 1$ for $t \geq 2T$, which implies

$$e_i(t_k) = e_i(s_i(t)) = \begin{cases} e^{i,1} & \text{if } t_k = 0, T \\ e^{i,2} & \text{if } t_k \geq 2T \end{cases}.$$

Given the agent's initial state, $q_i(0) = (3.2, 4.1, \pi/4, v_{max})$ and $\sigma_i(0) = \mathsf{ACC}$, its behavior is computed as follows. The configuration's evolution is

$$\begin{aligned} q_i(t) &= \phi_{f_i^*}(q_i(0), \sigma_i(0)) = \\ &= \left(3.2 + \tfrac{\sqrt{2}}{2} v_{max} t, 4.1 + \tfrac{\sqrt{2}}{2} v_{max} t, \pi/4, v_{max}\right), \end{aligned}$$

for $t < 2T$, since it also holds $\sigma_i(T) = \mathsf{ACC}$. Moreover, we have $\sigma_i(2T) = \mathsf{DEC}$ and hence

$$\begin{aligned} q_i(t) &= \phi_{f_i^*}(q_i(2T), \sigma_i(2T)) = \\ &= \left(3.2 + A, 4.1 + A, \pi/4, v_{max} e^{-\mu(t-2T)}\right), \end{aligned}$$

for $t \geq 2T$, where $A = \sqrt{2}\, v_{max} T + \bar{\Delta}(2T)$ and $\bar{\Delta}(t_k) = \tfrac{v_{max}}{\mu}\left(1 - e^{-\mu(t-t_k)}\right)$.

Consider another agent $\mathcal{A}_h$ trying to learn whether $\mathcal{A}_i$ is cooperative or not. Assume that $\mathcal{A}_h$ has only partial view of the region $\eta_{i,1}(q_i)$, i.e. $q_h$ is s.t. $\eta_{i,1}(q_i) \not\subseteq \mathcal{V}_h(q_1, \cdots, q_n)$, and thus $v_i = 0$. At $t = 0$, the local monitor reads the measures, $\bar{q}_i(0) = q_i(0)$ and $I_i^h(0) = \emptyset$, and initializes the estimate of $\mathcal{A}_i$'s states as

$$\begin{aligned} \hat{q}_i(0|0) &= \bar{q}_i(0) = q_i(0), \\ \hat{\sigma}_i(t_{-1}|0) &= \Sigma_i = \{\mathsf{ACC}, \mathsf{DEC}\}. \end{aligned}$$

Given that $\hat{s}_i(0|0) = 0$, the predicted behavior of the agent during the observation period $T_0$ is

$$\begin{aligned} \hat{\sigma}_i(0|0) &= \tilde{\delta}_i(\hat{\sigma}_i(t_{-1}|0), \tilde{e}_i(\hat{s}_i(0|0), v_i(0)) = \\ &= \tilde{\delta}_i(\{\mathsf{ACC}, \mathsf{DEC}\}, \{e^{i,1}, e^{i,2}\}) = \{\mathsf{ACC}, \mathsf{DEC}\}, \\ \hat{q}_i(t|0) &= \phi_{\hat{f}_i^*}(\hat{q}_i(0|0), \hat{\sigma}_i(0|0)) = \{q_i^{\mathsf{ACC}}(t|0), q_i^{\mathsf{DEC}}(t|0)\}, \end{aligned}$$

where

$$\begin{aligned} q_i^{\mathsf{ACC}}(t|0) &= \left(3.2 + \tfrac{\sqrt{2}}{2} v_{max} t, 4.1 + \tfrac{\sqrt{2}}{2} v_{max} t, \tfrac{\pi}{4}, v_{max}\right), \\ q_i^{\mathsf{DEC}}(t|0) &= \left(3.2 + \bar{\Delta}(0), 4.1 + \bar{\Delta}(0), \tfrac{\pi}{4}, v_{max} e^{-\mu t}\right). \end{aligned}$$

We also have $L(0) = \{(q_i^{\mathsf{ACC}}(t|0), \mathsf{ACC}), (q_i^{\mathsf{DEC}}(t|0), \mathsf{DEC})\}$. At $t = T$, the monitor reads the measures $\bar{q}_i(T) = q_i(T)$ and $I_i^h(T) = \emptyset$, and the agent's predicted state can be corrected as follows:

$$\begin{aligned} \hat{q}_i(T|T) &= \bar{q}_i(T), \\ \hat{\sigma}_i(0|T) &= \pi_{\Sigma_i}(L(0) \bowtie_{\varepsilon} q_i(t)) = \mathsf{ACC}, \end{aligned}$$

which gives, along with the fact that $\hat{\sigma}_i(0|0) \bowtie_{\tilde{\delta}_i} \hat{\sigma}_i(0|T) = e^{i,1}$, the a posteriori topology activation estimate $\hat{s}_i(0|T) = 0$. The estimate of the unknown topology activation is $\hat{p}_{i,1}(\hat{s}_i(0|0), \hat{s}_i(0|T)) = 0$, which means the agent's behavior is cooperative if, and only if, no other forklift is present also in the portion of its neighborhood that is out of the monitor's visibility, i.e. in $\eta_{i,1}(q_i) \setminus \mathcal{V}_h(q_1, \cdots, q_n)$. This finally gives the estimated neighbor configuration set $\hat{I}_i^h(0|T) = \emptyset$.

Similar computation is performed during the observation periods $T_1$ and $T_2$, which is omitted here for space reasons. At $t = 3T$, the local monitor reads the measures $\bar{q}_i(3T) = q_i(3T)$ and $I_i^h(3T) = \emptyset$, which gives the following a posteriori

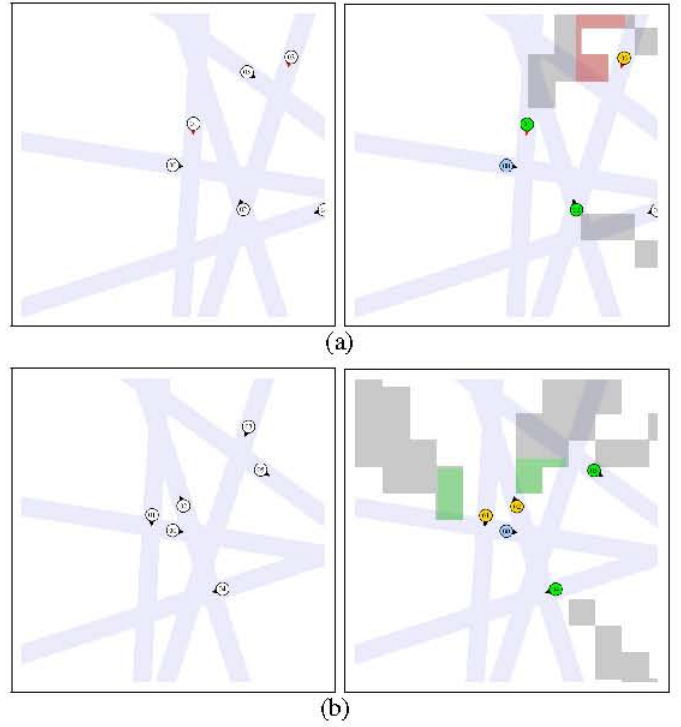

Figure 2. Simulation runs from the warehouse example with six cooperative forklifts: (left) state of the system; (right) neighborhood estimated by a local monitor onboard forklift 00.

estimates

$$\hat{q}_i(3T|3T) = \bar{q}_i(3T), \quad \hat{\sigma}_i(2T|3T) = \mathsf{DEC}.$$

Moreover, as $\hat{\sigma}_i(2T|2T) \bowtie_{\tilde{\delta}_i} \hat{\sigma}_i(2T|3T) = e^{i,2}$, we have $\hat{s}_i(2T|3T) = 1$ and $\hat{p}_{i,1}(\hat{s}_i(2T|2T), \hat{s}_i(2T|3T)) = 1$, which implies that the current behavior of forklift $\mathcal{A}_i$'s is compatible if, and only if, there is another forklift in the non–visible portion of $\eta_{i,1}(q_i)$. The estimated neighbor configuration set is indeed

$$\hat{I}_i^h(2T|3T) = \eta_{i,1}(q_i) \setminus \mathcal{V}_h(q_1, \cdots, q_n).$$

Note that an observer onboard the local monitor is unable to check if this estimated hypothesis is correct or not, which can be overcome as described in the following section.

Finally, consider the following two instants of a simulation with six cooperative forklifts. In the former, see Fig. 3–a, forklifts 00, 02, 04, and 05 are running in discrete state ACC and forklift 01 and 03 are in discrete state DEC. A local monitor onboard forklift 00 is trying to determine if its neighbors are cooperative or not. The monitor's decisions are the following: forklift 01 and 02 are cooperative; forklift 03's behavior is possibly cooperative (uncertain) and the presence of another forklift (05 in the figure) is inferred. In the latter instant, all cars are correctly running in discrete state ACC (Fig. 3–b). The figure shows that the local monitor is able to learn that forklift 04 and 05 are cooperative. Moreover, the behaviors of forklifts 01 and 02 are still uncertain, but the absence of other forklifts in the non–visible portion of $\eta_{1,1}(q_1)$ and $\eta_{2,1}(q_2)$ are correctly estimated.

## C. The Warehouse Example (Cont'd) - Corrupted Decoder

Consider an attack undertaken by means of alteration of the decoder map $u_i$ of an agent $\mathcal{A}_i$ described by a map of the following class:

$$\begin{aligned} u_i^\alpha \;:\; & \mathcal{Q} \times \Sigma_i \to \mathcal{U}_i \\ & (q_i, \text{ACC}) \mapsto (-\mu(1-\alpha)(v_i - v_{max}), 0)^T, \\ & (q_i, \text{DEC}) \mapsto (-\mu\, v_i, 0)^T, \end{aligned}$$

where $\alpha \in [0,1]$ is a fixed parameter. Alteration of the decoder map involves the agent's behavior only when the discrete state ACC is active. In particular, note that, for $\alpha \to 0$, the agent's "deviance" is almost negligible if $\alpha \to 0$ and may cause no damage to the system, whereas the case with $\alpha \to 1$ represents a more subtle type of attack, where the agent pretends to simulate the existence of another agent in $\eta_{i,1}(q_i)$ (the deviated motion is similar to that of a correct agent whose discrete state is DEC). The corresponding misbehavior model only differs from the nominal behavior in the controlled dynamic map

$$\dot{q}_i(t) = f_i(q_i(t), u_i^\alpha(q_i(t), \sigma_i(t_k))) = F_i^\alpha(q_i(t), \sigma_i(t_k)),$$

whose solution, $\phi_{F_i^\alpha}(q_i(t_k), \sigma_i(t_k))$, is immediately obtained by replacing $\mu$ with $\mu(1-\alpha)$ in Eq. 3 in the terms involving the state ACC, i.e.,

$$\begin{aligned} \Delta(\text{ACC}, t) &= v_{max}(t - t_k) + \tfrac{v_i(t_k) - v_{max}}{\mu(1-\alpha)}\left(1 - e^{-\mu(1-\alpha)(t-t_k)}\right), \\ V(\text{ACC}, t) &= v_{max}\left(1 - e^{-\mu(1-\alpha)(t-t_k)}\right). \end{aligned}$$

A suitable value for the monitor's accuracy $\epsilon$ can be computed based on the above misbehavior model, by guaranteeing that every deviated behavior is larger than the monitor's accuracy itself, i.e.,

$$\|\phi_{f_i^*}(q_i(t_k), \sigma) - \phi_{F_i^\alpha}(q_i(t_k), \sigma)\| > \epsilon, \; \forall \sigma \in \Sigma_i.$$

By assuming $\mu T$ small, this gives the conditions

$$\begin{aligned} \epsilon &\leq \|\phi_{f_i^*}(q_i(t_k), \text{ACC}) - \phi_{F_i^\alpha}(q_i(t_k), \text{ACC})\| \simeq \\ & \simeq \mu |v_i(t_k) - v_{max}| T \alpha, \\ \epsilon &\leq \|\phi_{f_i^*}(q_i(t_k), \text{DEC}) - \phi_{F_i^\alpha}(q_i(t_k), \text{ACC})\| \simeq \\ & \simeq -\mu |v_i(t_k) - v_{max}| T \alpha + \mu T v_{max} \end{aligned}$$

where the first–order Taylor series expansion of the exponential function, $e^x \simeq 1 + x$, was used. Finally, the monitor's accuracy must be greater than the minimum measurement precision $\epsilon_{min}^h$ of its onboard sensory system, i.e., $\epsilon \geq \epsilon_{min}^h$.

## D. The Highway Example (Cont'd) - Local Monitors

Consider four cars in the highway example (Fig. 4–a). Misbehavior of car 0, running a FAST maneuver along the second lane, while its next right lane is free, has to be detected (the car should start a RIGHT maneuver to return to the first lane). A FAST maneuver of a car in the second lane implies that the region on its right is occupied by another car. Three local monitors on the other cars try to learn whether the car 0 is cooperative or not, but have only partial view of the car's neighborhood. By means of the proposed local monitor, the three agents are able to compute estimates, $\hat{I}_0^1$, $\hat{I}_0^2$, and $\hat{I}_0^3$, of the occupancy map of car 0's neighborhood, which are reported in Fig. 4–b. However, the figure shows that all monitors are still unable to decide on the cooperativeness of the car 0, since there exist possible behaviors that comply with the cooperative model and their partial visibility.

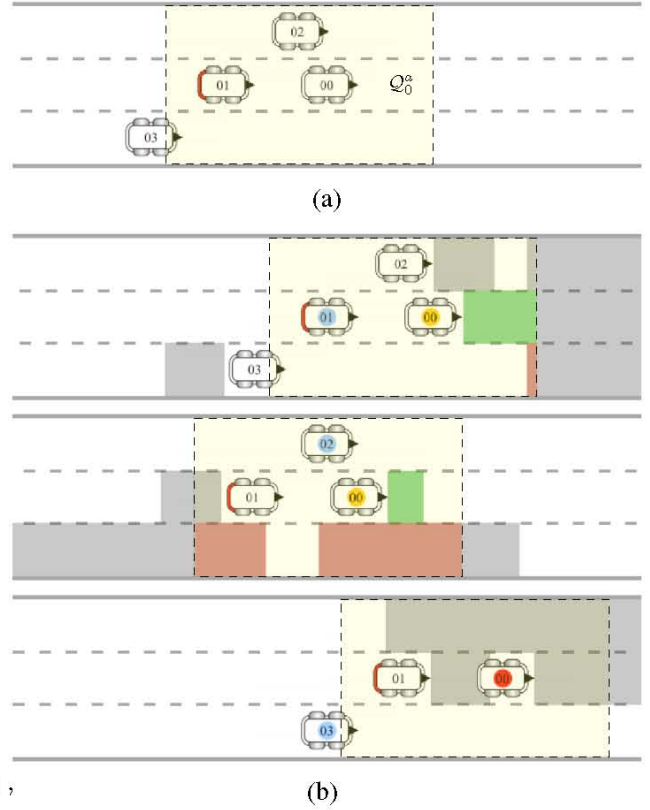

Figure 3. Misbehavior of car 0, running a FAST maneuver along the second lane, while its next right lane is free, has to be detected (a). Local maps of occupancy, $\hat{I}_0^1$, $\hat{I}_0^2$, and $\hat{I}_0^3$, which local monitors on the cars 1, 2, and 3 have reconstructed (b). The yellowish area dashed box outlines the target agent neighborhood; a blue circle specifies the current monitor; red (green) areas are non–visible regions, where the presence (absence) of a car is required. A colored circle around the target robot (green, yellow, or red) specifies its estimated cooperativeness (cooperative, uncertain, or uncooperative, respectively).

As a second example, consider eight cooperative cars in the highway and focus on the local view of car 0's monitor (Fig. 5). The presence of car 07 is detected (region $a$), based on the fact that car 06 is executing a SLOW maneuver. The presence of car 5 is detected (regions $e$, and $f$), based on the FAST maneuvers on the second lane executed by cars 3 and 4. This also allows the detection of car absence in front of car 3 (region $b$) and car 4 (region $c$). To the local monitor all these neighboring cars are uncertain, except car 1 that is certainly cooperative. The example is used to show the fact that — although this goes beyond the scope of the paper — a local monitor's uncertainty in the classification of a neighbor can be reduced by cross–correlating maps of occupancies of different neighbors: the occupancy map $\hat{I}_3^0$ contains a free region ($b$ in the figure) in front of car 3, and an occupied region (the union of $d$ with $e$) on its right, while $\hat{I}_2^0$ contains a free region (same $d$ in the figure) in front of it. Therefore the region $d$ in $\hat{I}_3^0$ must be removed and the only possibly occupied region must be ($e$).

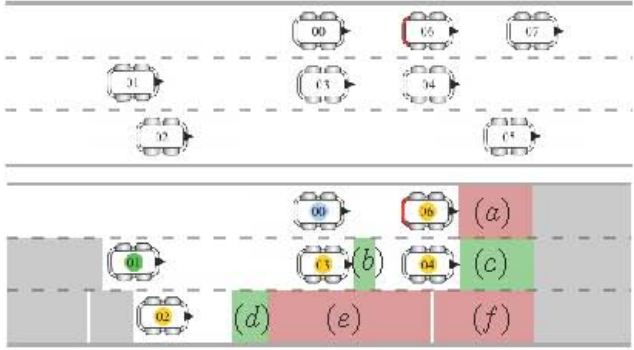

Figure 4. Eight cooperative cars (above) and view of the monitor on car 00 (below). A local monitor's uncertainty in the classification of a neighbor can be reduced by cross–correlating maps of occupancies of different neighbors.

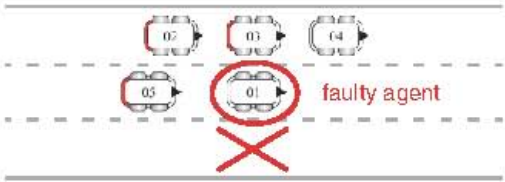

Figure 5. The misbehaving car 1 is executing a FAST maneuver on the second lane, while its next right lane is free.

### E. The Highway Example (Cont'd) - Monitor Agreement

Consider the example in Fig. 6 where four cars (2, 3, 4, and 5) are trying to reach a consensus on the misbehavior of a car (1 in the figure) remaining in the second lane. The cars can share their own local estimates of the occupancy map of car 1's neighborhood, by sending one–hop (immediate neighbor) messages through a communication network described by a connected graph $G = (V_G, E_G)$, with $V_G = \{2,3,4,5\}$ and $E_G = \{e_{2,2}, e_{2,3}, e_{2,5}, e_{3,3}, e_{3,4}, e_{4,4}, e_{5,5}\}$ (note that $\text{diam}(G) = 3$). The corresponding set–valued consensus protocol specializes to the following dynamic system:

$$\begin{cases} X_2(k+1) = F^{(3)}(X_2(k), X_3(k), X_5(k)) = \\ \qquad = X_2(k) \cap^+ X_3(k) \cap^+ X_5(k), \\ X_3(k+1) = F^{(3)}(X_2(k), X_3(k), X_4(k)) = \\ \qquad = X_2(k) \cap^+ X_3(k) \cap^+ X_4(k), \\ X_4(k+1) = F^{(2)}(X_3(k), X_4(k)) = X_3(k) \cap^+ X_4(k), \\ X_5(k+1) = F^{(2)}(X_2(k), X_5(k)) = X_2(k) \cap^+ X_5(k). \end{cases}$$

The system's evolution is reported in Fig. 7, where the $i$-th row represents the evolution of $X_i(t)$ (from left to right). No single local monitor has initially detected the misbehavior, which is instead iteratively obtained by car 2 and 3 after two consensus steps and then by the other two cars. As expected from theory, all local monitors consent to the centralized estimated occupancy map (last column in the figure)

$$X^* = \hat{I}_1 = F^{(4)}(\hat{I}_1^2, \hat{I}_1^3, \hat{I}_1^4, \hat{I}_1^5)$$

after at most 3 steps.

### VI. CONCLUSION

The problem of detecting misbehaving robots in a decentralized setting was addressed in this work. Robots are supposed to interact with each other based on "rules" that depends on the presence and absence of other agents in their neighborhoods. The literature on DES for fault diagnosability and state observability (see the pioneering work of Ramadge and Wohnam in [21] and successive of [24], [44]–[47]) addresses similar problems, but those solutions are not applicable mainly because the interaction topology in our systems is time–varying and unknown. Therefore, solutions should be valid for any topology. Furthermore, the common approach to fault diagnosis consists of exciting the model with suitable inputs and measuring the error between the expected output and the actually measured one [48], [49], requires changing the cooperation rules, whilst we are seeking a solution that is not invasive as such. The proposed solution is a distributed IDS where every agent first runs a local monitor to obtain a subjective map of free and occupied regions and then a consensus algorithm to agree on a unique shared map. The framework includes set–valued observers that allow local monitors to combine maps estimated at different time instants. In this paper, we have assumed that information exchanged is correct. Future extension of the work will address the important generalization to the case where robots can send false information due to communication failure or tampering. Preliminary but promising results in this direction are reported in [35].

### ACKNOWLEDGMENT

This work has been partially supported by the European Commission with contract FP7-IST-2008-224428 "CHAT - Control of Heterogeneous Automation Systems: Technologies for scalability, reconfigurability and security", with contract number FP7-2007-2-224053 CONET, the "Cooperating Objects Network of Excellence", with contract number FP7-2010-257649 PLANET, "PLAtform for the deployment and operation of heterogeneous NETworked cooperating objects", and with contract number FP7-2010-257462 HYCON2, "Highly-complex and networked control systems".

### APPENDIX

#### A. The Highway Example - Formalization

Consider $n$ cars that are supposed to follow the European, right–hand traffic rules while traveling along a highway with $m$ lanes. Every car $\mathcal{A}_i$ must coordinate its motion with neighboring cars as follows: accelerate up to its allowed maximum speed $v_{max}^i$ if the current lane is free; change to the next left lane if the current one is occupied by a preceding car and there are no cars on the immediate back; reduce speed and remain in the current lane otherwise; try to proceed along the next right lane when possible; do not overtake on the right.

The system can be described as an instance of $\mathcal{P}$ with the environment, the configuration $q_i$ and the dynamic map $f_i : \mathcal{Q} \times \Sigma_i \to T_\mathcal{Q}$ described in Section II-A. Moreover, we need to introduce a topology $\eta_{i,1}(q_i)$ representing a region in the immediate front of the agent, a topology $\eta_{i,2}(q_i)$ for a region on its left, a topology $\eta_{i,3}(q_i)$ for a region on its right, and a

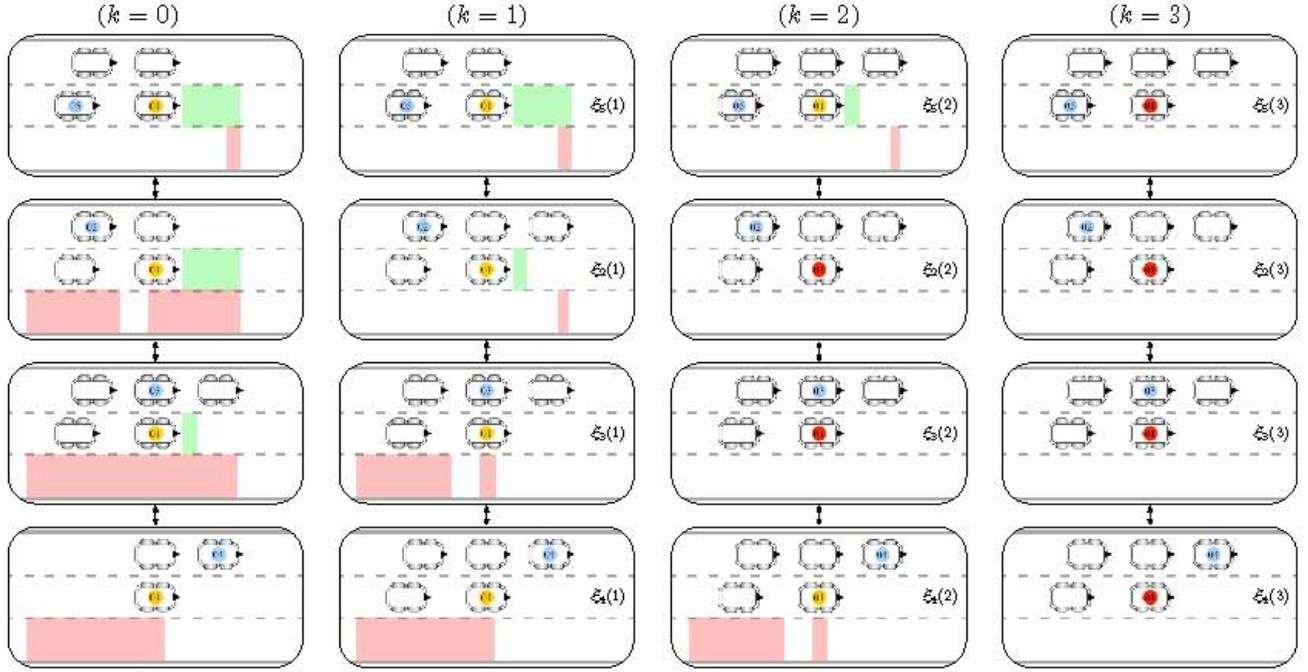

Figure 6. Misbehavior of car 1 is detected by the set–valued consensus algorithm, although no single local monitor was initially able to do it.

topology $\eta_{i,4}(q_i)$ for a region on its back (Fig. 2). These are formally described as

$$\eta_{i,1} : \mathcal{Q} \to 2^{\mathcal{Q}}$$
$$q_i \mapsto \{(x,y,\theta,v) \,|\, x_i \le x \le x_i + d_f,$$
$$\lfloor \tfrac{y_i}{w} \rfloor w \le y \le (\lfloor \tfrac{y_i}{w} \rfloor + 1)\,w\},$$

$$\eta_{i,2} : \mathcal{Q} \to 2^{\mathcal{Q}}$$
$$q_i \mapsto \{(x,y,\theta,v) \,|\, x_i - d_b \le x \le x_i + d_f,$$
$$(\lfloor \tfrac{y_i}{w} \rfloor + 1)\,w \le y \le (\lfloor \tfrac{y_i}{w} \rfloor + 2)\,w\},$$

$$\eta_{i,3} : \mathcal{Q} \to 2^{\mathcal{Q}}$$
$$q_i \mapsto \{(x,y,\theta,v) \,|\, x_i - d_b \le x \le x_i + d_f,$$
$$(\lfloor \tfrac{y_i}{w} \rfloor - 1) \le y \le \lfloor \tfrac{y_i}{w} \rfloor w\},$$

$$\eta_{i,4} : \mathcal{Q} \to 2^{\mathcal{Q}}$$
$$q_i \mapsto \{(x,y,\theta,v) \,|\, x_i - d_b \le x \le x_i,$$
$$\lfloor \tfrac{y_i}{w} \rfloor w \le y \le (\lfloor \tfrac{y_i}{w} \rfloor + 1)\,w\},$$

where $w$ is the lane width, $d_f$ and $d_b$ are a forward and backward safety distances, and $\lfloor \cdot \rfloor$ returns the nearest lower integer of the argument. Thus, the encoder map is $s_i : \mathcal{Q} \times \mathcal{Q}^{n_i} \to \mathbb{B}^4$, $s_i = (s_{i,1}, \cdots, s_{i,4})$, and the agent's neighborhood is $N(q_i) = \eta_{i,1}(q_i) \cup \cdots \cup \eta_{i,4}(q_i)$. Moreover, we need to introduce two constants $\lambda_{i,1}, \lambda_{i,2}$ representing the left–most and right–most lanes, respectively, and two constants $\lambda_{i,3}, \lambda_{i,4}$ representing the current target lane's left and right edges, respectively:

$$\lambda_{i,1} = \{(x,y,\theta,v) \,|\, (m-1)w \le y \le mw\},$$
$$\lambda_{i,2} = \{(x,y,\theta,v) \,|\, 0 \le y \le w\},$$
$$\lambda_{i,3} = \{(x,y,\theta,v) \,|\, y = \left(\lfloor \tfrac{y_i(t_k)}{w} \rfloor + 1\right) w\},$$
$$\lambda_{i,4} = \{(x,y,\theta,v) \,|\, y = \lfloor \tfrac{y_i(t_k)}{w} \rfloor w\}.$$

The event alphabet is $E_i = \{e^{i,1}, \cdots, e^{i,13}\}$ and the detector map $e_i \in \mathbb{B} \to 2^{E_i}$ is characterized by the event conditions

$$c_{i,1} = \neg s_{i,1} s_{i,3},\quad c_{i,2} = \neg s_{i,1}\lambda_{i,2},\quad c_{i,3} = s_{i,1} s_{i,2},$$
$$c_{i,4} = s_{i,1} s_{i,4},\quad c_{i,5} = s_{i,1}\lambda_{i,1},\quad c_{i,6} = s_{i,1}\neg s_{i,2}\neg s_{i,4}\neg\lambda_{i,1},$$
$$c_{i,7} = \neg s_{i,1}\neg s_{i,3}\neg\lambda_{i,2},\quad c_{i,8} = \neg s_{i,1},\quad c_{i,9} = \lambda_{i,3},$$
$$c_{i,10} = s_{i,1}\neg\lambda_{i,3},\quad c_{i,11} = s_{i,1},\quad c_{i,12} = \lambda_{i,4},$$
$$c_{i,13} = \neg s_{i,1}\neg\lambda_{i,4}.$$

The finite set of discrete states is $\Sigma_i = \{\text{FAST}, \text{SLOW}, \text{LEFT}, \text{RIGHT}\}$ ($p = 4$) and the automaton's dynamics is

$$\delta_i : \Sigma_i \times 2^{E_i} \to \Sigma_i$$
$$(\text{FAST}, e^{i,1}), (\text{FAST}, e^{i,2}) \mapsto \text{FAST},$$
$$(\text{FAST}, e^{i,3}), (\text{FAST}, e^{i,4}), (\text{FAST}, e^{i,5}) \mapsto \text{SLOW},$$
$$(\text{FAST}, e^{i,6}) \mapsto \text{LEFT},$$
$$(\text{FAST}, e^{i,7}) \mapsto \text{RIGHT},$$
$$(\text{SLOW}, e^{i,8}) \mapsto \text{FAST},$$
$$(\text{SLOW}, e^{i,3}), (\text{SLOW}, e^{i,4}), (\text{SLOW}, e^{i,5}) \mapsto \text{SLOW},$$
$$(\text{SLOW}, e^{i,6}) \mapsto \text{LEFT},$$
$$(\text{LEFT}, e^{i,8}), (\text{LEFT}, e^{i,9}) \mapsto \text{FAST},$$
$$(\text{LEFT}, e^{i,10}) \mapsto \text{LEFT},$$
$$(\text{RIGHT}, e^{i,11}), (\text{LEFT}, e^{i,12}) \mapsto \text{FAST},$$
$$(\text{RIGHT}, e^{i,13}) \mapsto \text{RIGHT},$$

with initial state $\sigma_i^0 = \text{FAST}$.

The decoder map is $u_i : \mathcal{Q} \times \Sigma_i \to \mathcal{U}_i$, $u_i = (a_i, \omega_i)$, with

$$a_i : \mathcal{Q} \times \Sigma_i \to \mathbb{R}$$
$$(q_i, \text{FAST}), (q_i, \text{LEFT}), (q_i, \text{RIGHT}) \mapsto \begin{cases} \bar{a} & \text{if } v_i < v_{max}^i \\ 0 & \text{otherwise} \end{cases},$$
$$(q_i, \text{SLOW}) \mapsto \begin{cases} -\bar{a} & \text{if } v_i > 0 \\ 0 & \text{otherwise} \end{cases},$$

$$\omega_i : \mathcal{Q} \times \Sigma_i \to \mathbb{R}$$
$$(q_i, \text{FAST}), \ (q_i, \text{SLOW}) \mapsto \left((y^*(q_i) - y_i)\frac{\sin\theta_i}{\theta_i} - \mu\theta_i\right)v_i,$$
$$(q_i, \text{LEFT}) \mapsto \begin{cases} \bar{\omega} & \text{if } \theta_i < \theta_{max} \\ 0 & \text{otherwise} \end{cases},$$
$$(q_i, \text{RIGHT}) \mapsto \begin{cases} -\bar{\omega} & \text{if } \theta_i > -\theta_{max} \\ 0 & \text{otherwise} \end{cases},$$

where $y^*(q_i) = \left(\lfloor \frac{y_i}{w} \rfloor + \frac{1}{2}\right)w$ is the current lane center, $\theta_{max}$ is the agent's maximum curvature angle, and $\mu$, $\bar{a}$ and $\bar{\omega}$ are positive constants.

Finally, the visibility map returns the set of configurations laying within a distance $R_i$ and that are not hidden by other cars (see e.g. the known sweeping line algorithm in [50] for its computation, and the examples in Fig. 8). A formal description of the map is avoided for space reasons.

### B. Event Estimation with Incomplete, Time–Varying Visibility

A proof of the formula used for the observer's detector map of Eq. 4 is given in this section. This result, along with the procedure presented in Section III for the construction of the nondeterministic automaton $\tilde{\delta}_i$ extends available solutions (see e.g. [39]) in sofar as that it shows that an observer for discrete event systems with uncertain events can be efficiently estimated also with incomplete, time–varying visibility.

First consider the following propositions:

*Proposition 1:* The smallest upper approximation of a detector condition $c_{i,j} = s_{i,k}$ ($\gamma_{i,j} = \{k\}, \rho_{i,j} = \mu_{i,j} = \pi_{i,j} = \emptyset$), based on an observer's topology check $v_h$ and an available encoder map $\tilde{s}_{i,k}$, is

$$\tilde{c}_{i,j} = \tilde{s}_{i,k}\, v_{h,k} + \neg v_{h,k}\,.$$

*Proof:* Based on the observer's visibility region $V_h$, the encoder map $s_{i,k}$ can be written as

$$s_{i,k}(q_i, I_i) = \tilde{s}_{i,k}(q_i, I_i^h) + \sum_{q_k \in I_i \setminus V_h} \mathbf{1}_{\eta_{i,k}(q_i)}(q_k) =$$
$$= \tilde{s}_{i,k}(q_i, I_i^h) + \tilde{p}_{i,k}(q_i, I_i)\,,$$

that can be conveniently factorized as follows. If $\tilde{p}_{i,k} = 0$, the expression reduces to $c_{i,j} = \tilde{s}_{i,k}$, whereas if $\tilde{p}_{i,k} = 1$, it becomes $c_{i,j} = \tilde{s}_{i,k} + 1 = 1$. Then, the detector condition can be factorized as $c_{i,j} = \tilde{s}_{i,k}\,\neg \tilde{p}_{i,k} + 1\,\tilde{p}_{i,k}$. Moreover, if the observer has complete visibility of the $k$-th topology ($v_{i,k} = 1$), $\tilde{p}_{i,k} = 0$ since $I_i \setminus V_h = \emptyset$, which implies $c_{i,j} = \tilde{s}_{i,k}$, whereas nothing can be said on the value of $\tilde{p}_{i,k}$ if $v_{i,k} = 0$. Therefore, $c_{i,j}$ can be factorized w.r.t. the observer's topology check as

$$c_{i,j} = \tilde{s}_{i,k}\, v_{i,k} + (\tilde{s}_{i,k}\,\neg \tilde{p}_{i,k} + \tilde{p}_{i,k})\,\neg v_{i,k}\,.$$

Its visibility-based smallest upper approximation is

$$\tilde{c}_{i,j} = \max_{\tilde{p}_{i,k} \in \mathbb{B}} c_{i,j} = \tilde{s}_{i,k}\, v_{i,k} + A\,\neg v_{i,k}\,,$$

with $A = \max_{\tilde{p}_{i,k} \in \mathbb{B}}\left(\tilde{s}_{i,k}\,\neg \tilde{p}_{i,k} + \tilde{p}_{i,k}\right) = \max\{\tilde{s}_{i,k}, 1\} = 1$, which proves the thesis. ■

*Proposition 2:* The smallest upper approximation of a detector condition $c_{i,j} = \neg s_{i,k}$ ($\gamma_{i,j} = \emptyset, \rho_{i,j} = \{k\}, \mu_{i,j} = \pi_{i,j} = \emptyset$), based on an observer's topology check $v_h$ and an available encoder map $\tilde{s}_{i,k}$, is

$$\tilde{c}_{i,j} = \neg \tilde{s}_{i,k}\,.$$

*Proof:* As in Prop. 1, based on the observer's visibility region $V_h$, the detector condition $c_{i,j}$ can be written as

$$\neg s_{i,k}(q_i, I_i) = \neg\left(\tilde{s}_{i,k}(q_i, I_i^h) + \tilde{p}_{i,k}(q_i, I_i)\right) =$$
$$= \neg\tilde{s}_{i,k}(q_i, I_i^h)\,\neg\tilde{p}_{i,k}(q_i, I_i)\,,$$

where De Morgan's law is used. If $\tilde{p}_{i,k} = 0$, the expression reduces to $c_{i,j} = \neg\tilde{s}_{i,k}$, whereas if $\tilde{p}_{i,k} = 1$, it becomes $c_{i,j} = 0$. Then, $c_{i,j}$ can be factorized as $c_{i,j} = \neg\tilde{s}_{i,k}\,\neg\tilde{p}_{i,k} + 0\,\tilde{p}_{i,k} = \neg\tilde{s}_{i,k}\,\neg\tilde{p}_{i,k}$. Moreover, if $v_{i,k} = 1$, $\tilde{p}_{i,k} = 0$ that implies $c_{i,j} = \neg\tilde{s}_{i,k}$, whereas nothing can be said on its value otherwise. Therefore, $c_{i,j}$ can be factorized w.r.t. the observer's topology check as

$$c_{i,j} = \neg\tilde{s}_{i,k}\, v_{i,k} + \neg\tilde{s}_{i,k}\,\neg\tilde{p}_{i,k}\,\neg v_{i,k}$$

Its visibility-based smallest upper approximation is

$$\tilde{c}_{i,j} = \neg\tilde{s}_{i,k}\, v_{i,k} + \neg\tilde{s}_{i,k}\,\max_{\tilde{p}_{i,k} \in \mathbb{B}}(\tilde{p}_{i,k})\,\neg v_{i,k} =$$
$$= \neg\tilde{s}_{i,k}\, v_{i,k} + \neg\tilde{s}_{i,k}\,\neg v_{i,k} =$$
$$= \neg\tilde{s}_{i,k}\,(v_{i,k} + \neg v_{i,k}) = \neg\tilde{s}_{i,k}\,,$$

which gives the thesis. ■

*Proposition 3:* The smallest upper approximation of a detector condition $c_{i,j} = s_{i,k}\,\neg s_{i,m}$ ($\gamma_{i,j} = \{k\}, \rho_{i,j} = \{m\}, \mu_{i,j} = \pi_{i,j} = \emptyset$), based on an observer's topology check $v_h$, is

$$\tilde{c}_{i,j} = (\tilde{s}_{i,k}\, v_{h,k} + \neg v_{h,k})\,\neg\tilde{s}_{i,m}\,.$$

*Proof:* Based on the observer's visibility region $V_h$, the detector condition can be written as

$$c_{i,j} = (\tilde{s}_{i,k} + \tilde{p}_{i,k})(\neg\tilde{s}_{i,m}\,\neg\tilde{p}_{i,m}) =$$
$$= \tilde{s}_{i,k}\,\neg\tilde{s}_{i,m}\,\neg\tilde{p}_{i,m} + \tilde{p}_{i,k}\,\neg\tilde{s}_{i,m}\,\neg\tilde{p}_{i,m}\,.$$

By enumerating all possible combinations of $\tilde{p}_{i,k}$ and $\tilde{p}_{i,m}$, $c_{i,j}$ can be factorized as

$$c_{i,j} = (\neg\tilde{s}_{i,m})\,\tilde{p}_{i,k}\,\neg\tilde{p}_{i,m} + (\tilde{s}_{i,k}\,\neg\tilde{s}_{i,m})\,\neg\tilde{p}_{i,k}\,\neg\tilde{p}_{i,m}\,.$$

Moreover, based on the observer's topology check (recall that $v_{i,k} = 1$ implies $\tilde{p}_{i,k} = 0$, and $v_{i,m} = 1$ implies $\tilde{p}_{i,m} = 0$), the expression can be further factorized as

$$c_{i,j} = A\, v_{i,k}\, v_{i,m} + B\, v_{i,k}\,\neg v_{i,m} +$$
$$+ C\,\neg v_{i,k}\, v_{i,m} + D\,\neg v_{i,k}\,\neg v_{i,m}\,,$$

with $A = \tilde{s}_{i,k}\,\neg\tilde{s}_{i,m}$, $B = \tilde{s}_{i,k}\,\neg\tilde{s}_{i,m}\,\neg\tilde{p}_{i,m}$, $C = \neg\tilde{s}_{i,m}\,\tilde{p}_{i,k} + (\tilde{s}_{i,k}\,\neg\tilde{s}_{i,m})\,\neg\tilde{p}_{i,k}$, and $D = \neg\tilde{s}_{i,m}\,\tilde{p}_{i,k}\,\neg\tilde{p}_{i,m} + \tilde{s}_{i,k}\,\neg\tilde{s}_{i,m}\,\neg\tilde{p}_{i,k}\,\neg\tilde{p}_{i,m}$. Its visibility-based smallest upper approximation is

$$\tilde{c}_{i,j} = \tilde{s}_{i,k}\,\neg\tilde{s}_{i,m}\,(v_{i,k}\, v_{i,m} + v_{i,k}\,\neg v_{i,m}) +$$
$$+ \neg\tilde{s}_{i,m}\,(\neg v_{i,k}\, v_{i,m} + \neg v_{i,k}\,\neg v_{i,m}) =$$
$$= \tilde{s}_{i,k}\,\neg\tilde{s}_{i,m}\, v_{i,k} + \neg\tilde{s}_{i,m}\,\neg v_{i,k}\,,$$

which easily gives the thesis. ■

We can now readily give a proof of Theorem 1 as follows. W.r.t. the above propositions, an event estimator map $e_i$ with detector conditions of the form of Eq. 1 is characterized by a generic combination of the sets $\gamma_{i,j}, \rho_{i,j} \in \{1, \cdots, \kappa_i\}$ and $\mu_{i,j}, \pi_{i,j} \in \{1, \cdots, h_i\}$. It is sufficient to show that the above propositions also extend to the general case.

*Proof: (of Theorem 1)* Let us proceed by induction. Consider the case with only $\gamma_{i,j} \neq \emptyset$ and $\text{card}(\gamma_{i,j}) \geq 1$. Assume $\gamma_{i,j} = \{1, \cdots, l\}$, which is always possible upon reordering of the encoder map's components. The case with $l=1$ is proved by Prop. 1. By assuming that the thesis holds for $l=m$, i.e., that the smallest upper approximation of $c_{i,j} = \Pi_{k \in \gamma_{i,j}} s_{i,k} = \Pi_{k=1}^{m} s_{i,k}$ is $\tilde{c}_{i,j} = (\Pi_{k=1}^{m} \tilde{s}_{i,k} v_{i,k} + \neg v_{i,k})$, the inductive step requires proving it for $l=m+1$. Indeed, the detector condition $c_{i,j} = \Pi_{k=1}^{m+1} s_{i,k}$ can be written as

$$\underbrace{\left(\Pi_{m=1}^{k} s_{i,k}\right)}_{z} s_{i,m+1} = z\, s_{i,m+1} = z\,(\tilde{s}_{i,m+1} + \tilde{p}_{i,m+1})\,,$$

that can be factorized as follows. If $\tilde{p}_{i,m+1}=0$, the expression reduces to $c_{i,j} = z\,\tilde{s}_{i,m+1}$, whereas if $\tilde{p}_{i,m+1}=1$, it becomes $c_{i,j}=z$, thus giving the expression

$$c_{i,j} = z\,\tilde{s}_{i,m+1}\,\neg \tilde{p}_{i,m+1} + z\,\tilde{p}_{i,m+1}\,.$$

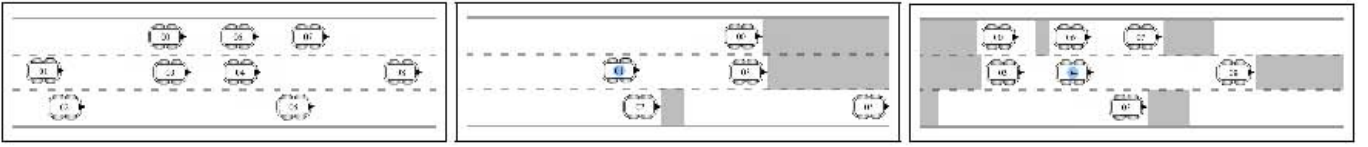

Figure 7. Sensing model in the highway example: from left to right, complete state of the system and views of agents $\mathcal{A}_1$ and $\mathcal{A}_4$, respectively.

The detector condition can be factorized w.r.t. the observer's topology check $v_{i,m+1}$ as follows. If $v_{i,m+1}=1$, we have $\tilde{p}_{i,m+1}=0$ and $c_{i,j}=z\,\tilde{s}_{i,m+1}$, whereas if $v_{i,m+1}=0$ nothing can be said on its value. This yields

$$c_{i,j} = z\,A\,v_{i,m+1} + z\,B\,\neg v_{i,m+1}\,,$$

with $A = \tilde{s}_{i,m+1}$, and $B = \tilde{s}_{i,m+1}\,\neg \tilde{p}_{i,m+1} + \tilde{p}_{i,m+1}$. Its visibility–based, smallest upper approximation is

$$\tilde{c}_{i,j} = \max_{\tilde{p}_{i,1},\dots,\tilde{p}_{i,m+1} \in \mathcal{B}} c_{i,j} =$$
$$= \max_{\tilde{p}_{i,1},\dots,\tilde{p}_{i,m}} z \cdot \max_{\tilde{p}_{i,m+1}} (A\,v_{i,m+1} + B\,\neg v_{i,m+1}) =$$
$$= \left(\Pi_{k=1}^{m} \tilde{s}_{i,k} v_{i,k} + \neg v_{i,k}\right) \left(\tilde{s}_{i,m+1} v_{i,m+1} + \neg v_{i,m+1}\right),$$

which proves the thesis in the first considered case.

Consider the case with only $\rho_{i,j} \neq \emptyset$, $\rho_{i,j} = \{1,\dots,l\}$. As above, we want to proceed by induction. The case with $l=1$ is proved by Prop. 2. By assuming that the thesis holds for $l=m$, i.e., that the smallest upper approximation of $c_{i,j} = \Pi_{k \in \rho_{i,j}} \neg s_{i,k} = \Pi_{k=1}^{m} \neg s_{i,k}$ is $\tilde{c}_{i,j} = \Pi_{k=1}^{m} \neg \tilde{s}_{i,k}$, the inductive step requires proving it for $l=m+1$. Indeed, the detector condition

$$c_{i,j} = \Pi_{k=1}^{m+1} \neg s_{i,k} = z\,\neg \tilde{s}_{i,m+1}\,\neg \tilde{p}_{i,m+1}$$

can be factorized as follows. If $\tilde{p}_{i,m+1}=0$, the expression reduces to $c_{i,j} = z\,\neg \tilde{s}_{i,m+1}$, whereas, if $\tilde{p}_{i,m+1}=1$, it becomes $c_{i,j}=0$, thus giving the expression

$$c_{i,j} = z\,\neg \tilde{s}_{i,m+1}\,\neg \tilde{p}_{i,m+1}\,.$$

The detector condition can be factorized w.r.t. the observer's topology check $v_{i,m+1}$ as follows. If $v_{i,m+1}=1$, we have $\tilde{p}_{i,m+1}=0$ and $c_{i,j} = z\,\neg \tilde{s}_{i,m+1}$, whereas if $v_{i,m+1}=0$ nothing can be said on its value. This yields

$$c_{i,j} = z\,\neg \tilde{s}_{i,m+1} v_{i,m+1} + z\,\neg \tilde{s}_{i,m+1}\,\neg \tilde{p}_{i,m+1}\,\neg v_{i,m+1} =$$
$$= z\,\neg \tilde{s}_{i,m+1}\,(v_{i,m+1} + \tilde{p}_{i,m+1}\,\neg v_{i,m+1})\,.$$

Its visibility–based, smallest upper approximation is
$\tilde{c}_{i,j} = (\Pi_{k=1}^{m} \tilde{s}_{i,k} v_{i,k} + \neg v_{i,k})\,\neg \tilde{s}_{i,m+1} C$, with

$$C = \max_{\tilde{p}_{i,m+1}} (v_{i,m+1} + \neg \tilde{p}_{i,m+1}\,\neg v_{i,m+1}) =$$
$$= \max\{v_{i,m+1}, 1\} = 1\,,$$

which proves the thesis also in this second case.

The cases with $\gamma_{i,j}, \rho_{i,j} \neq \emptyset$ and their cardinality greater than the unity straightforwardly follow from the discussion above and recursive application of Prop. 3. Finally, the estimated value of every application $\lambda_{i,k}$, affecting $c_{i,j}$ if $\mu_{i,j}, \tau_{i,j} \neq \emptyset$, coincides with its real value as they only depend on the configuration $q_i$ of the monitored agent $\mathcal{A}_i$, that is measurable from $\mathcal{A}_h$ by assumption. ∎